\documentclass[10pt,twocolumn,letterpaper]{article}

\usepackage{cvpr}              
\usepackage{graphicx}
\usepackage{multicol}
\usepackage{tabularx}
\usepackage{multirow}
\usepackage{color, colortbl}
\usepackage{xcolor}
\usepackage[normalem]{ulem}
\usepackage{float}
\usepackage{mdframed}
\usepackage{comment}
\usepackage{booktabs}

\usepackage{amssymb}

\usepackage{pifont}
\newcommand{\cmark}{\ding{51}}
\newcommand{\xmark}{\ding{55}}

\definecolor{darkpastelgreen}{rgb}{0.01, 0.75, 0.24}
\definecolor{lightgray}{gray}{0.9}

\newcommand{\keypoint}[1]{\noindent\textbf{#1}\quad}

\newcommand{\dataset}{Ground-V}
\definecolor{cvprblue}{rgb}{0.21,0.49,0.74}
\usepackage[pagebackref,breaklinks,colorlinks,allcolors=cvprblue]{hyperref}

\hypersetup{
  colorlinks = true, 
  urlcolor = blue, 
  linkcolor = red, 
  citecolor = blue 
}
\def\paperID{6650} 
\def\confName{CVPR}
\def\confYear{2025}

\title{\dataset: Teaching VLMs to Ground Complex Instructions in Pixels
}

\author{Yongshuo Zong$^{1}$\thanks{Work done during internship at Amazon.}, \ Qin Zhang$^2$\thanks{Corresponding author: qzaamz@amazon.com}, \ Dongsheng An$^2$, \ Zhihua Li$^2$, \ Xiang Xu$^2$, \  Linghan Xu$^2$, \\
Zhuowen Tu$^2$, \ Yifan Xing$^2$, \ Onkar Dabeer$^2$ \\ [.5ex]
$^1$ University of Edinburgh \qquad $^2$ AWS AI Labs  \\
{\tt\footnotesize yongshuo.zong@ed.ac.uk, \{qzaamz,andongsh,zhihuaa,xiangx,linghanx,ztu,yifax,onkardab\}@amazon.com}
}

\begin{document}
\maketitle
\begin{abstract}
This work presents a simple yet effective workflow for automatically scaling instruction-following data to elicit pixel-level grounding capabilities of VLMs under complex instructions. In particular, we address five critical real-world challenges in text-instruction-based grounding: hallucinated references, multi-object scenarios, reasoning, multi-granularity, and part-level references. By leveraging knowledge distillation from a pre-trained teacher model, our approach generates high-quality instruction-response pairs linked to existing pixel-level annotations, minimizing the need for costly human annotation. The resulting dataset, \textbf{\emph{\dataset}}, captures rich object localization knowledge and nuanced pixel-level referring expressions. Experiment results show that models trained on {\dataset} exhibit substantial improvements across diverse grounding tasks. 
Specifically, incorporating \dataset~during training directly achieve an average accuracy boost of 4.4\% for LISA and a 7.9\% for PSALM across six benchmarks on the gIoU metric. 
It also sets new state-of-the-art results on standard benchmarks such as RefCOCO/+/g. Notably, on gRefCOCO, we achieve an N-Acc of 83.3\%, exceeding the previous state-of-the-art by more than 20\%. 
\end{abstract}

\section{Introduction}
\label{sec:intro}
Picture this: you are at a fruit bar and someone asks you to ``\textit{add more fruits with high antioxidants}.'' Without hesitation, your mind automatically translates this abstract request into action -- your eyes scanning for and pinpointing apples, pomegranates, or blueberries. This seemingly simple process demonstrates an important human cognitive ability: connecting prior knowledge with visual recognition, navigating multiple levels of abstraction, and precisely localizing objects based on imprecise instructions. While humans perform such tasks naturally, current state-of-the-art computer vision specialist~\cite{zou2024seem,yan2023uninext,liu2024groundingdinomarryingdino,ren2024grounded}, despite their excellent performance on standard object detection or segmentation benchmarks, struggle to interpret abstract natural language instructions for precise visual grounding, particularly when handling complex instructions and visual scenarios.

\begin{figure}[t]
  \centering
  \includegraphics[width=1\columnwidth]{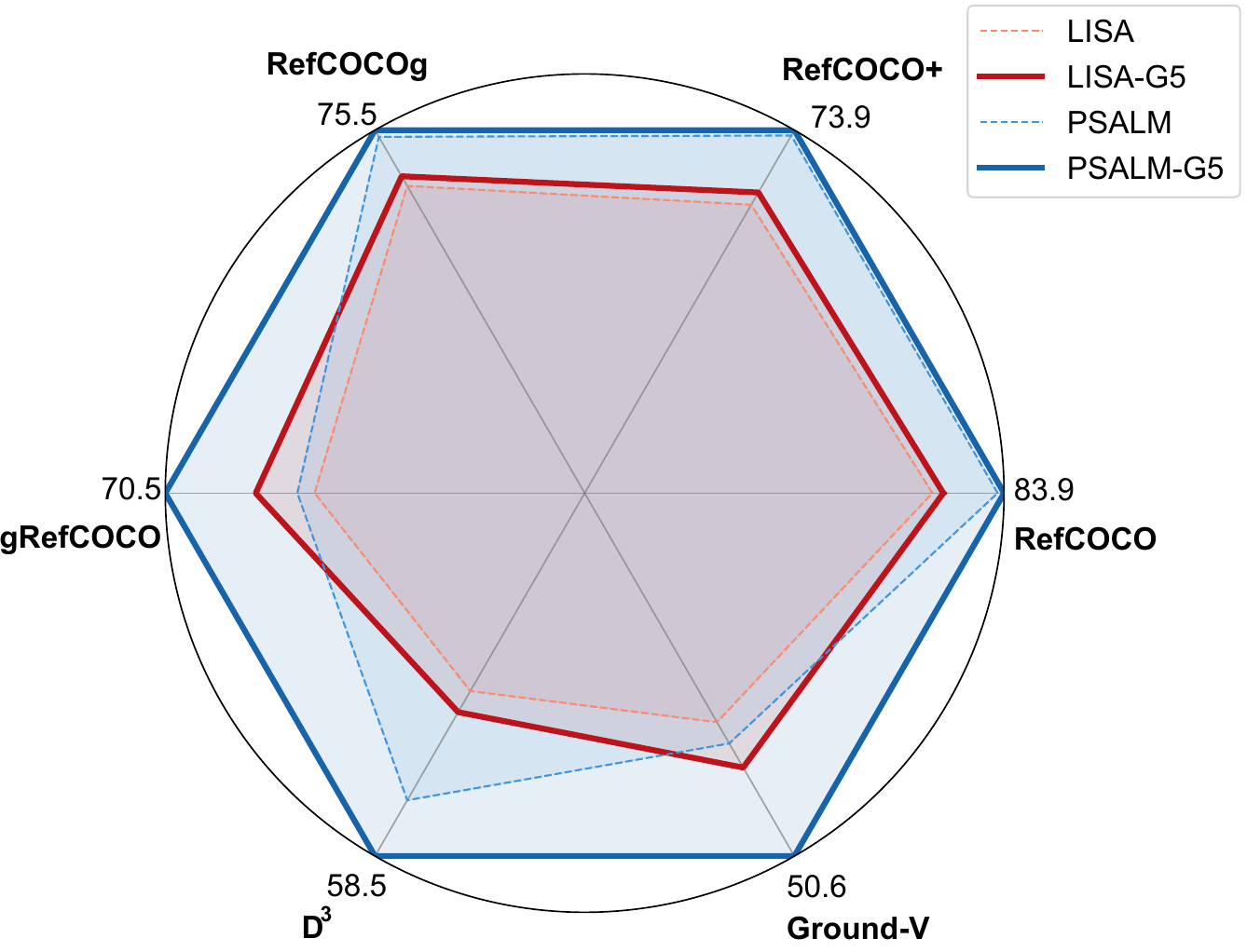}
  \caption{Performance comparison of LISA and PSALM models w/ and w/o our \dataset~dataset during training. Incorporating \dataset \ consistently enhances both models' performance across benchmarks, achieving an average improvement of 4.4\% for LISA and 7.9\% for PSALM on the gIoU metric across six benchmarks.
  }
  \label{fig:teaser}
\end{figure}

Advances in large vision-language models (VLMs)~\citep{liu2023llava, bai2023qwen, dai2023instructblip, zhu2024minigpt, liu2024improved} have shown promise toward bridging this gap. These models exhibit strong capabilities in general multimodal tasks such as recognition, perception, and multimodal reasoning. To further enhance VLMs' ability to not only recognize and understand images but also localize specific regions of interest in pixel level, i.e. segmentation, researchers have proposed architectural designs to incorporate the strengths of computer vision specialists. By learning special grounding tokens in VLMs~\citep{lai2024lisa, ren2024pixellm, xia2024gsva, zhang2024psalm} and using them as inputs for computer vision specialists like SAM~\citep{kirillov2023sam}, these models are beginning to unlock emerging capabilities such as reasoning-driven segmentation~\citep{lai2024lisa}.


However, we find that current VLMs for segmentation are often unreliable under complex instructions or challenging visual scenarios\footnote{We provide detailed visualizations of this issue in the appendix.}. For example, in an image with multiple red, green, and yellow apples, when asked to segment only the red apples, these VLMs often produce inaccurate results by including apples of other colors. This problem becomes more pronounced with complex instructions, such as ``\emph{segment the partially eaten red apple next to the ceramic bowl}'', where VLMs typically ignore contextual details, segmenting random ``\emph{red apples}'' without considering whether they are partially eaten or near the specified bowl. We conjecture that one root cause of these limitations in the current VLM training data used for grounding tasks: while humans naturally describe visual scenes using rich, nuanced language, most grounding datasets contain only simple, direct references. This disconnect between human-like descriptive complexity and oversimplified training data limits these VLMs' ability to translate the understanding of complex instructions into precise localization in the real world.




Inspired by the success of data scaling in language instruction-following~\citep{chung2024scaling, wang2023selfinstruct}, we ask a natural question: \textit{can scaling of pixel-grounded instruction following data bridge this gap?} To explore this, we scale such data by incorporating richer scenarios and increasing the number of samples. We develop an automated data generation workflow with minimal human annotation (only for the test data) to tackle five critical challenges in real-world referring segmentation: (1) multi-object interactions, (2) multi-granular instructions, (3) hallucinated references, (4) reasoning, and (5) part references. Using this workflow, we introduce \textbf{\emph{\dataset}} (where ``V'' denotes the Roman numeral for five)\footnote{For simplicity, we use the acronym ``G5'' interchangeably.}, a comprehensive dataset of 500K instruction-segmentation pairs. We find that training with \dataset~effectively connects the multimodal understanding capabilities of VLMs with the grounding expertise of segmentation specialists, eliciting pixel-level grounding under complex text instructions. Importantly, our dataset can be seamlessly integrated into the training of existing models, yielding immediate performance improvements, as shown in Figure~\ref{fig:teaser}. Experiment results demonstrate that models trained with \emph{\dataset} not only achieve state-of-the-art performance on standard benchmarks (e.g., RefCOCO-series~\citep{yu2016refcoco, nagaraja2016modeling} and gRefCOCO~\citep{liu2023gres}) but also show substantial improvements of up to 20\% on our more challenging test set. 
In summary, we make the following contributions:

\begin{itemize}
    \item We develop an automated data generation workflow with minimal human input to address five critical challenges in real-world referring segmentation. 
    To our knowledge, we are the first work to systematically consider these challenging scenarios and enable scalable generation of pixel-grounded instruction-following data.
    \item Using the proposed data generation workflow, we introduce \emph{\dataset}, a comprehensive dataset of 500K instruction-segmentation pairs for visual grounding. The large-scale training set is crafted to enhance model performance in complex scenarios. We also provide a detailed evaluation protocol with human-annotated test data to accurately assess model performance.
    \item We conduct thorough evaluations on both public datasets and the human-annotated test set of \emph{\dataset}, achieving new state-of-the-art results. Specifically, we achieve 70.6\% gIoU and 83.7\% N-Acc on gRefCOCO on average, surpassing previous state-of-the-art by 5.7\% in gIoU and 20.9\% in N-Acc. On the more challenging \dataset~test set, we achieve an average boost of 15.4\% in gIoU over the state-of-the-art PSALM model~\citep{zhang2024psalm}.
\end{itemize}

\section{Related Work}
\label{sec:related}
\noindent\textbf{Specialist Models for Referring Segmentation} \ Recent advances in object segmentation have led to several powerful state-of-the-art models, such as Mask2Former~\citep{cheng2022masked},  SegNext~\citep{guo2022segnext}, and SAM~\citep{kirillov2023sam, ravi2024sam2segmentimages}, which exhibit strong generalization across different scenes and object classes. Despite their benchmark-leading performance, these models lack a nuanced understanding of and interaction with natural language. This limitation is apparent in referring segmentation tasks~\citep{yu2016refcoco, nagaraja2016modeling}, where the objective is to segment target objects based on descriptive language instructions. There have been strong interest in developing specialists for referring segmentation~\citep{ding2021vision, zhu2022seqtr, ye2019cross, yu2018mattnet, wang2022cris, yang2022lavt}. Recent advances include PolyFormer~\citep{liu2023polyformer}, which frames the task as sequential polygon generation through a sequence-to-sequence framework; UNINEXT~\citep{yan2023uninext}, which incorporates text inputs to unify instance-perception tasks; and SEEM~\citep{zou2024seem}, which encodes user intents into prompts within a joint visual-semantic space. Nevertheless, most of these methods focus on architectural unification and multi-task compatibility -- leaving the critical challenge of handling complex language instructions in real-world scenarios unaddressed.

\begin{figure*}[t!]
  \centering
\includegraphics[width=0.99\textwidth]{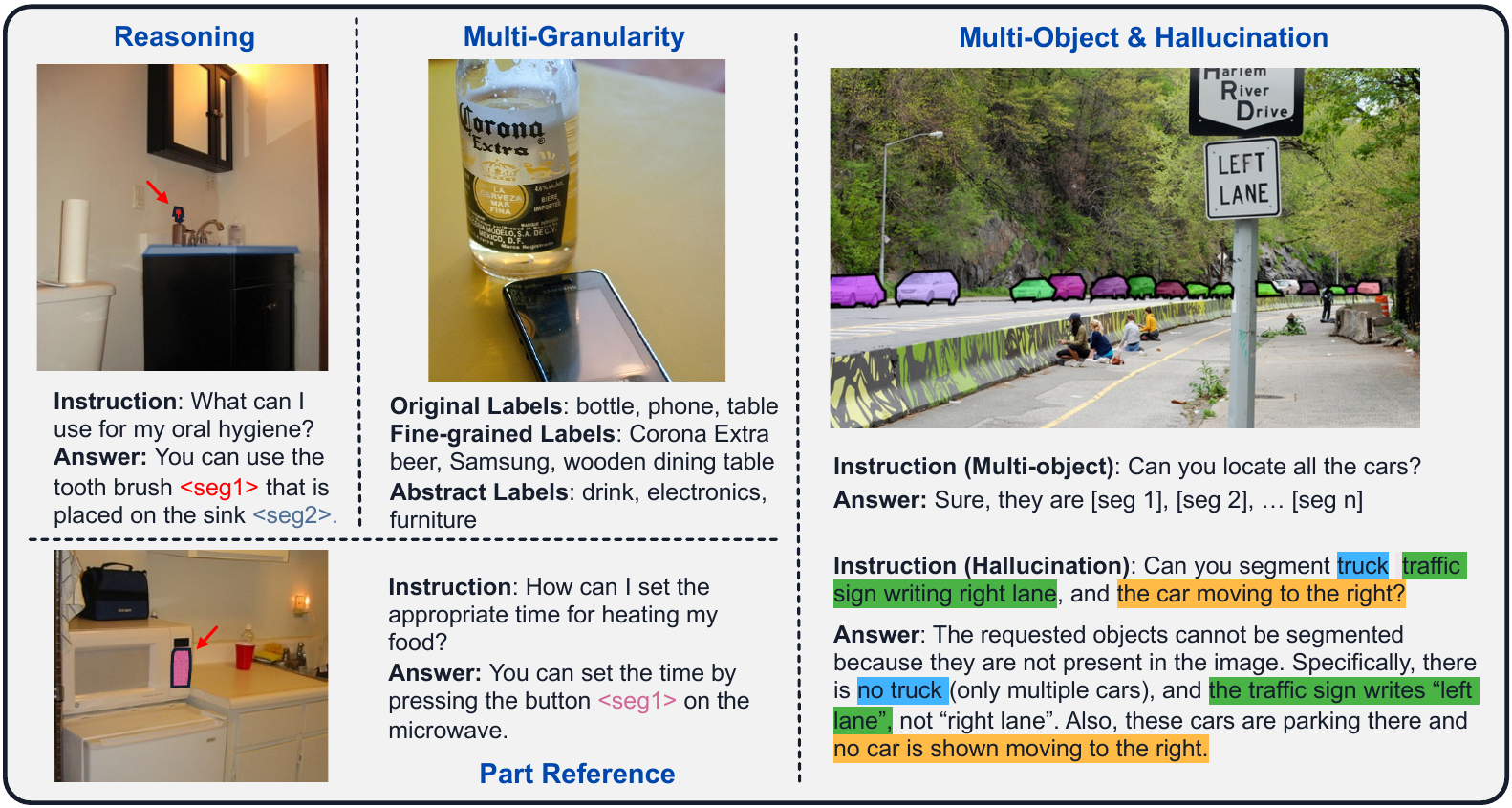}
  \vspace{-3pt}
  \caption{Illustration of the diverse scenarios covered in our \dataset~dataset. These include reasoning-based segmentation (top left), multi-granular instructions (center), multi-object and hallucination handling (top right), and part-whole relationships (bottom left). Each example demonstrates how our dataset provides rich, nuanced instructions and corresponding segmentations.
  }
  \label{fig:data_example}
\end{figure*}

\noindent\textbf{Generalist Vision-Language Models} \
Large VLMs extend the capabilities of LLMs by incorporating visual components -- typically through vision encoders and/or decoders connected via specialized stitching mechanisms~\citep{li2023blip, liu2024improved, liu2023llava, bai2023qwen, zong2024self, peng2024grounding, zhang2023internlmx}. These models have made rapid advancements alongside LLMs and have garnered significant attention for their impressive multi-modal capabilities in tasks such as recognition, captioning, reasoning, and visual question answering. However, most generalist VLMs are limited to text-based outputs, typically lacking built-in capabilities for vision-centric tasks like segmentation and detection. Recent studies~\citep{tong2024eyes,rahmanzadehgervi2024vision, ranasinghe2024learning} have also identified various visual shortcomings in these generalist VLMs, such as weak spatial reasoning and localization awareness, highlighting the need for enhanced visual processing capabilities.

\noindent\textbf{VLMs for Segmentation} \
Several recent works have tried to equip VLMs with the ability to generate segmentation masks. LISA~\citep{lai2024lisa} is a pioneering approach that introduces a learnable segmentation token within VLMs, using it as a prompt embedding for the SAM decoder to predict segmentation masks. Building on LISA, u-LLaVA~\citep{xu2023u} extends support to object grounding tasks. However, LISA is limited to single-object scenarios, prompting later works like GLaMM~\citep{rasheed2024glamm}, PerceptionGPT~\citep{pi2024perceptiongpt}, PixelLM~\citep{zhang2024psalm}, GSVA~\citep{xia2024gsva}, and PSALM~\citep{zhang2024psalm} to address multi-object scenarios through various learnable condition tokens. Despite strong performance on standard benchmarks, we observe that these models become highly unreliable when handling more complex instructions. We hypothesize that this limitation arises from the models’ potential being constrained by the limited richness of their training data, despite being designed specifically for referring segmentation.

\begin{figure*}[t!]
  \centering
  \includegraphics[width=0.98\textwidth]{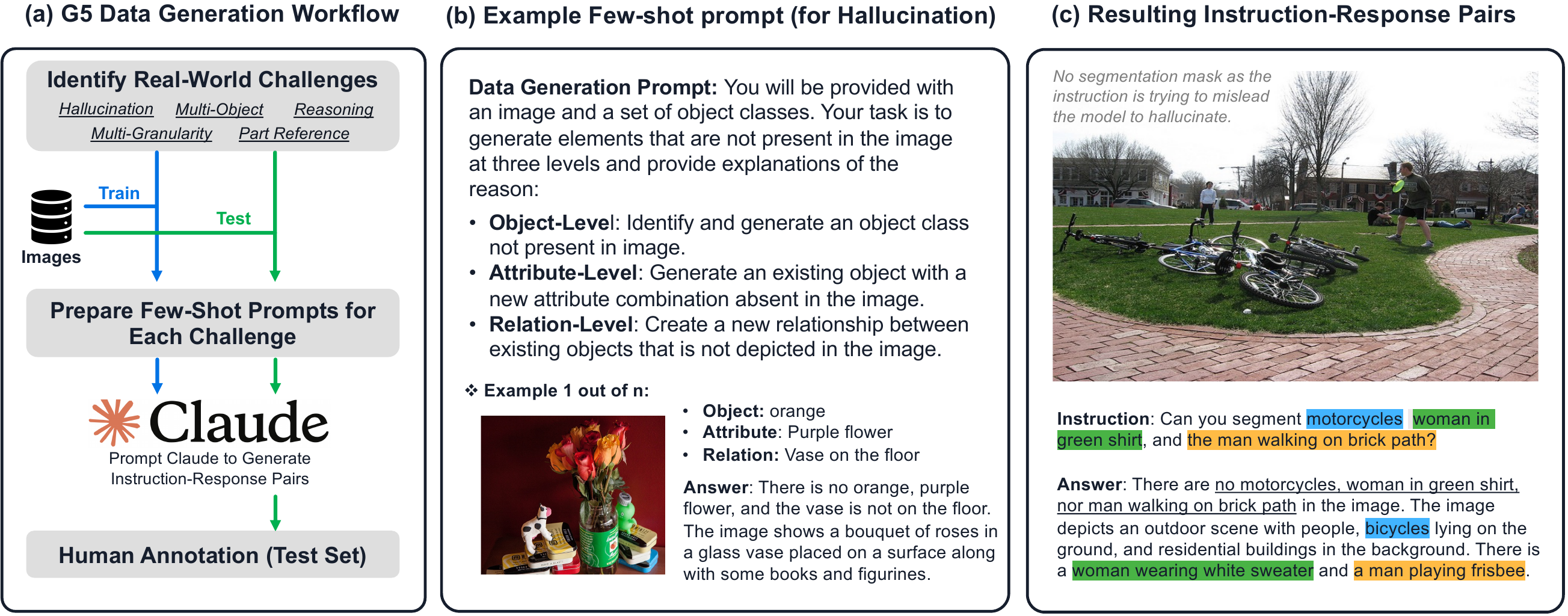}
  \vspace{-3pt}
  \caption{(a) Overview of the G5 data generation pipeline. We identify key real-world challenges, including hallucination, multi-object scenarios, reasoning, multi-granularity, and part-level reference. For each dimension, we design few-shot prompts to generate instruction-response pairs. Using these prompts alongside corresponding images, we guide Claude to produce instruction-response pairs. Finally, human annotators validate the evaluation set to ensure accuracy and reliability. (b) Here we give an example few-shot prompt for generating hallucination-mitigation data, covering three levels: object-level, attribute-level, and relation-level. (c) We display a sample of the generated instructions and responses, paired with segmentation masks from the original data source, which supports training and evaluation. For hallucination-mitigation data, however, no segmentation mask is included because the instruction is designed to mislead the model deliberately; a correctly functioning model should not produce a segmentation mask in this scenario.}
  
  \label{fig:data}
\end{figure*}

\section{\dataset}
This section outlines five key dimensions essential for effective reasoning-based segmentation in real-world scenarios: multi-granular instructions, multi-object handling, hallucinated references, part references, and reasoning-driven segmentation. We propose a scalable approach to curate the training and test sets in \dataset, incorporating these five dimensions, as shown in Figure~\ref{fig:data} and Table~\ref{tab:dataset_comparison}.

\subsection{Five Real-World Challenges}
\noindent\textbf{Multi-granularity} \
Objects in the real world can be described at various levels of conceptual specificity. For instance, a corgi could be referred to as a specific breed (``corgi"), a general category (``dog"), a functional role (``pet"), or a broad biological classification (``animal"). Humans tend to use higher-level, abstract references when uncertain about specific details. 
To capture this variability in referring specificity, we implement a hierarchical label structure incorporating both \textit{abstract} and \textit{fine-grained} annotations. An abstract label can cover multiple objects, while fine-grained distinctions can be made within the same class.

\noindent\textbf{Multi-object} \ In addition to handling single-object referring expressions at multiple levels of granularity, we also address the simultaneous segmentation of multiple objects by introducing instructions that reference more than five objects. This task aligns with the generalized referring expression setting~\citep{liu2023gres} but presents a greater challenge that requires the VLM to handle quantity effectively.

\begin{table*}[t]
\centering
\resizebox{0.975\linewidth}{!}{
\begin{tabular}{lcccccc}
\toprule
\textbf{Datasets} & \textbf{Reasoning} & \textbf{Multi-Object} & \textbf{Hallucination} & \textbf{Part Reference} & 
\textbf{Multi-Granularity} & \textbf{\# of Instruction pairs}  \\
\midrule
ReasonSeg & \textcolor{darkpastelgreen}{\cmark} & \textcolor{red}{\xmark} & \textcolor{red}{\xmark} & \textcolor{red}{\xmark} & \textcolor{red}{\xmark} & 1.2K \\ 
MUSE      & \textcolor{darkpastelgreen}{\cmark} & \textcolor{darkpastelgreen}{\cmark} & \textcolor{red}{\xmark} & \textcolor{red}{\xmark} & \textcolor{red}{\xmark} & 214K \\ 
\textbf{\dataset \ \small{(acronym ``G5'')}} & \textcolor{darkpastelgreen}{\cmark} & \textcolor{darkpastelgreen}{\cmark} & \textcolor{darkpastelgreen}{\cmark} & \textcolor{darkpastelgreen}{\cmark} & \textcolor{darkpastelgreen}{\cmark} & 481K \\ 
\bottomrule
\end{tabular}
}
\vspace{-3pt}
\caption{Comparison between our proposed {\dataset} and existing datasets designed for VLM segmentation, including ReasonSeg~\cite{lai2024lisa} and MUSE~\citep{ren2024pixellm}. Our dataset surpasses previous ones in both comprehensiveness and scale. 
}
\label{tab:dataset_comparison}
\end{table*}


\begin{table*}[t!]
\centering
\resizebox{0.975\linewidth}{!}{
\begin{tabular}{lcccp{0.03cm}ccc}
\toprule
& \multicolumn{3}{c}{\textbf{Train Set}} & & \multicolumn{3}{c}{\textbf{Test Set \footnotesize{(Human Annotated)}}}  \\
\cmidrule{2-4}\cmidrule{6-8}
\textbf{Subset} & \textbf{\# Images} & \textbf{\# Instructions} & \textbf{Avg. Objects / Instr.} & &\textbf{\# Images} & \textbf{\# Instructions} & \textbf{Avg. Objects / Instr.} \\
\midrule
Multi-Object & 28,902   & \phantom{0}36,850 &  10.01 & & 1,251 &  \phantom{0}1,585   & 9.96  \\
Hallucination & 21,019  &  \phantom{0}63,057  & \phantom{0}7.21 & & 5,000 & 15,000 & 7.36 \\
Reasoning & 41,624 &  \phantom{0}41,624 & \phantom{0}6.45 &  & 4,817 & \phantom{0}4,817 & 6.72  \\
Part Reference & 41,807  & 105,908   & \phantom{0}1.94 & &  2,036 &  10,391   &   1.89  \\
Multi-Granularity (Coarse) & 35,268  &  \phantom{0}80,498 & \phantom{0}3.15 & & 5,000 & 11,536 & 3.19 \\
Multi-Granularity (Fine) &  30,041  & \phantom{0}95,878  & \phantom{0}1.91 & & 4,282  & 14,262  & 1.88\\
\midrule
Total &  50,000 &  423,815  & \phantom{0}4.09 & & 5,000 & 57,591 & 4.20 \\
\bottomrule
\end{tabular}
}
\vspace{-3pt}
\caption{Statistics of the \dataset~train and test sets.}
\label{tab:statistics}
\end{table*}

\noindent\textbf{Part Reference} \ We aim to enable VLMs to ground finer object parts for compositional localization.  For instance, in the Figure~\ref{fig:data_example} microwave example, if a robot needs to use the microwave, it must not only locate the microwave but also identify and segment its button. To support this, we incorporate part annotations from the PACO dataset~\citep{ramanathan2023paco} and rephrase the labels into natural language instructions.

\noindent\textbf{Reasoning} \ A key advantage of using VLMs for grounding tasks is their ability to leverage reasoning to interpret and act on abstract instructions that are nuanced and require contextual understanding (e.g., ``steak'' vs. ``food with more protein''). In such cases, the model should be able to comprehend the underlying context and produce appropriate segmentation masks. Building on the reasoning segmentation setup in~\citep{lai2024lisa}, we extend this framework by allowing the model to handle instructions that involve either multiple objects or a single object, in contrast to ReasonSeg~\citep{lai2024lisa}, which is limited to single-object responses.

\noindent\textbf{Hallucinated Reference} \ To prevent potential hallucinations in real-world segmentation tasks, such as segmenting objects that are not present in the image despite being instructed to do so, we introduce three types of no-target scenarios to evaluate the model's robustness against hallucinated references: object, attribute, and relation. Here, \textit{object hallucinations} refers to items that are absent from the image, \textit{attribute hallucinations} pertains to properties not associated with the objects, and \textit{relation hallucinations} involves nonexistent relationships between objects, such as incorrect spatial positioning. We assign empty segmentation masks to these inputs and provide text descriptions that explain why they should be excluded from segmentation.

\subsection{\dataset \ Data Generation Workflow}
We develop an automatic and scalable approach to generate instruction-segmentation pairs (we also provide bounding boxes), along with potential text answers for each sample. The overall pipeline is illustrated in Figure~\ref{fig:data}.

\noindent\textbf{Detailed Data Generation Process} \ For each of the five aforementioned dimensions, we manually craft a 3-shot demonstration, with each shot consisting of an image, an instruction, and a corresponding answer. These examples serve as in-context demonstrations~\citep{brown2020gpt3} for querying a teacher VLM, Claude 3 Sonnet\footnote{https://www.anthropic.com/news/claude-3-family}, which then generates new instruction-response pairs based on the query images. This process is applied to generate both the training and evaluation sets. For the evaluation set, we conduct a second round of re-validation using Claude 3.5 Sonnet as a judge model, regenerating instruction-answer pairs when the original outputs are deemed incorrect. Finally, human annotators manually inspect the whole evaluation set to ensure correctness. We use COCO 2017~\citep{lin2014mscoco} train and validation split as our image source as well as segmentation masks and bounding boxes as annotations for training and evaluation sets. During the data generation, Claude is prompted to link the instructions and text answers with corresponding object annotations (e.g., segmentation masks). In total, we have curated 50,000 images for training set and 5,000 images for evaluation set. The detailed statistics are shown in Table~\ref{tab:statistics}. We provide some exemplar data in Figure~\ref{fig:data_example}. More details such as words frequency, category distribution, and detailed prompts can be found in the supplementary materials.

\begin{table*}[t!]
\centering
\begin{tabularx}{0.975\textwidth}{ll*{3}{>{\centering\arraybackslash}X}p{0.02cm}*{3}{>{\centering\arraybackslash}X}p{0.02cm}*{2}{>{\centering\arraybackslash}X}p{0.02cm}c}
\toprule
& \multirow{2}{*}{\textbf{Method}} & \multicolumn{3}{c}{\textbf{RefCOCO}} & & \multicolumn{3}{c}{\textbf{RefCOCO+}} & & \multicolumn{2}{c}{\textbf{RefCOCOg}} & & \multirow{2}{*}{\textbf{Average}} \\
\cmidrule{3-5} \cmidrule{7-9} \cmidrule{11-12}
 & & \textbf{val} & \textbf{testA} & \textbf{testB} & & \textbf{val} & \textbf{testA} & \textbf{testB} & & \textbf{val} & \textbf{test} &  \\
\midrule
\multirow{4}{*}{\textcolor{gray}{\rotatebox[origin=c]{90}{Specialist}}} 
 & SEEM-L & - & - & - & & - & - & - & & 65.6 & - & & - \\
 & PolyFormer-L & 76.9 & 78.5 & 74.8 & & 72.2 & 75.7 & 66.7 & & 71.2 & 71.2 & & 73.4 \\
 & UNINEXT-L & 80.3 & 82.6 & 77.8 & & 70.0 & 74.9 & 62.6 & & 73.4 & 73.7 & & 74.4 \\
 & UNINEXT-H & 82.2 & 83.4 & 81.3 & & 72.5 & 76.4 & 66.2 & & 74.7 & \textbf{76.4} & & 76.6 \\
\midrule
\multirow{9}{*}{\textcolor{gray}{\rotatebox[origin=c]{90}{VLM-Based}}} 
 & GLaMM & 79.5 & 83.2 & 76.9 & & 72.6 & \textbf{78.7} & 64.6 & & 74.2 & 74.9 & & 75.6 \\
 & u-LLava & 80.4 & 82.7 & 77.8 & & 72.2 & 76.6 & 66.8 & & 74.8 & 75.6 & & 75.9 \\
 & PerceptionGPT & 75.3 & 79.1 & 72.1 & & 68.9 & 74.0 & 61.9 & & 70.7 & 71.9 & & 71.7 \\
 & PixelLM & 73.0 & 76.5 & 68.2 & & 66.3 & 71.7 & 58.3 & & 69.3 & 70.5 & & 69.2 \\
 & GSVA & 77.7 & 79.9 & 74.2 & & 68.0 & 71.5 & 61.5 & & 73.2 & 73.9 & & 72.5 \\
 & LISA & 70.2 & 73.4 & 67.7 & & 59.2 & 63.9 & 53.0 & & 63.2 & 64.7 & & 64.4 \\
 & \cellcolor{lightgray}{LISA-G5} & \cellcolor{lightgray}{73.9} & \cellcolor{lightgray}{75.5} & \cellcolor{lightgray}{68.4} & \cellcolor{lightgray}& \cellcolor{lightgray}{63.1} & \cellcolor{lightgray}{65.7} & \cellcolor{lightgray}{54.7} & \cellcolor{lightgray}& \cellcolor{lightgray}{64.9} & \cellcolor{lightgray}{66.9} &\cellcolor{lightgray} & \cellcolor{lightgray}{66.6}  \\
 & PSALM & 83.6 & 84.7 & 81.6 & & 72.9 & 75.5 & \textbf{70.1} & & 73.8 & 74.4 & & 77.1 \\
 & \cellcolor{lightgray}{PSALM-G5} & \cellcolor{lightgray}{\textbf{83.9}} & \cellcolor{lightgray}{\textbf{85.0}} & \cellcolor{lightgray}{\textbf{82.1}} & \cellcolor{lightgray}& \cellcolor{lightgray}{\textbf{73.1}} & \cellcolor{lightgray}{75.8} & \cellcolor{lightgray}{69.8} & \cellcolor{lightgray}& \cellcolor{lightgray}{\textbf{74.8}} & \cellcolor{lightgray}{74.2} & \cellcolor{lightgray}& \cellcolor{lightgray}{\textbf{77.3}} \\
\bottomrule
\end{tabularx}
\vspace{-3pt}
\caption{Comparison with the state-of-the-art methods on three referring image segmentation benchmarks (RefCOCO, RefCOCO+, and RefCOCOg) using the cIoU metric. Models shaded in \smash{\colorbox{lightgray}{light gray}} are trained with the inclusion of our proposed \dataset. Our models achieve the best performance on most of the subsets.}
\label{tab:refcoco}
\end{table*}

\begin{table*}[t!]
\centering
\resizebox{0.975\linewidth}{!}{
\begin{tabular}{lllllp{0.01cm}lllp{0.01cm}lllp{0.01cm}lll}
\toprule
& \multirow{2}{*}{\textbf{Method}} & \multicolumn{3}{c}{\textbf{val}} & & \multicolumn{3}{c}{\textbf{testA}} & & \multicolumn{3}{c}{\textbf{testB}} & & \multicolumn{3}{c}{\textbf{Average}} \\
\cmidrule{3-5} \cmidrule{7-9} \cmidrule{11-13} \cmidrule{15-17}
& & \textbf{gIoU} & \textbf{cIoU} & \textbf{N-Acc} & & \textbf{gIoU} & \textbf{cIoU} & \textbf{N-Acc} & & \textbf{gIoU} & \textbf{cIoU} & \textbf{N-Acc} & & \textbf{gIoU} & \textbf{cIoU} & \textbf{N-Acc} \\
\midrule
\multirow{3}{*}{\textcolor{gray}{\rotatebox[origin=c]{90}{Specialist}}} &
 MattNet & 48.2 & 47.5 & 41.2 & & 59.3 & 58.7 & 44.0 & & 46.1 & 45.3 & 41.3 & & 51.2 & 50.5 & 42.2 \\
& LTS & 52.7 & 52.3 & - & & 62.6 & 61.9 & - & & 50.4 & 49.9 & - & & - & - & - \\
& LAVT & 58.4 & 57.6 & 49.3 & & 65.9 & 65.3 & 49.3 & & 55.8 & 55.0 & 48.5  & & 60.0 & 59.3 & 49.0 \\
& ReLA & 63.6 & 62.4 & 56.4 & & 70.0 & 69.3 & 59.0 & & 61.0 & 59.9 & 58.4 & & 64.9 & 63.9 & 57.9 \\
\midrule
\multirow{10}{*}{\textcolor{gray}{\rotatebox[origin=c]{90}{VLM-based}}}
& GSVA & 63.3 & 61.7 & 62.4 & & 70.1 & 69.6 & 65.3 & & 61.3 & 60.3 & 60.6 & & 64.9 & 63.9 & 62.8 \\
& LISA & 32.2 & 38.7 & \phantom{0}2.7 & & 48.5 & 52.6 & \phantom{0}6.3 & & 39.7 & 44.8 & \phantom{0}5.0 & & 40.1 & 45.4 & \phantom{0}4.7 \\
& \cellcolor{lightgray}{LISA-G5} & \cellcolor{lightgray}{46.7} & \cellcolor{lightgray}{48.4} & \cellcolor{lightgray}{36.4} & \cellcolor{lightgray}& \cellcolor{lightgray}{63.2} & \cellcolor{lightgray}{63.7} & \cellcolor{lightgray}{39.9} & \cellcolor{lightgray}& \cellcolor{lightgray}{51.3} & \cellcolor{lightgray}{53.9} & \cellcolor{lightgray}{44.1} & \cellcolor{lightgray}& \cellcolor{lightgray}{53.7} & \cellcolor{lightgray}{55.3} & \cellcolor{lightgray}{40.1} \\
& PSALM & 43.3 & 42.0 & 27.7 & & 54.5 & 52.4 & 20.3 & & 52.5 & 50.6 & 25.6 & & 50.1 & 48.3 & 24.5 \\
& \cellcolor{lightgray}{PSALM-G5} & \cellcolor{lightgray}{\textbf{64.6}} & \cellcolor{lightgray}{\textbf{67.3}} & \cellcolor{lightgray}{\textbf{83.3}} & \cellcolor{lightgray}& \cellcolor{lightgray}{\textbf{74.5}} & \cellcolor{lightgray}{\textbf{71.8}} & \cellcolor{lightgray}{\textbf{83.2}} & \cellcolor{lightgray}& \cellcolor{lightgray}{\textbf{72.7}} & \cellcolor{lightgray}{\textbf{72.3}} & \cellcolor{lightgray}{\textbf{84.6}} & \cellcolor{lightgray}& \cellcolor{lightgray}{\textbf{70.6}} & \cellcolor{lightgray}{\textbf{70.5}} & \cellcolor{lightgray}{\textbf{83.7}} \\
\cmidrule{2-16}
& GSVA (ft) & 66.5 & 66.4 & 66.0 & & 71.1 & 72.8 & 64.7 & & 62.2 & 63.2 & 62.5 & & 66.6 & 67.5 & 64.4 \\
& LISA (ft) & 63.3 & 61.7 & 56.5 & & 70.1 & 69.2 & 63.5 & & 61.3 & 60.3 & 58.4 & & 64.9 & 63.7 & 59.5 \\
& \cellcolor{lightgray}{LISA-G5 (ft)} & \cellcolor{lightgray}{68.2} & \cellcolor{lightgray}{65.6} & \cellcolor{lightgray}{61.3} &\cellcolor{lightgray} & \cellcolor{lightgray}{74.6} & \cellcolor{lightgray}{72.6} & \cellcolor{lightgray}{66.9} & \cellcolor{lightgray}& \cellcolor{lightgray}{66.3} & \cellcolor{lightgray}{65.0} & \cellcolor{lightgray}{63.2} & \cellcolor{lightgray}& \cellcolor{lightgray}{69.7} & \cellcolor{lightgray}{67.7} & \cellcolor{lightgray}{63.8} \\
& PSALM (ft) & - & \textbf{69.3} & 78.3 & & - & 74.8 & 73.4 & & - & 71.3 & 79.6 & & - & 71.8 & 77.1 \\
& \cellcolor{lightgray}{PSALM-G5 (ft)} & \cellcolor{lightgray}{\textbf{67.3}} & \cellcolor{lightgray}{68.0} & \cellcolor{lightgray}{\textbf{87.2}} & \cellcolor{lightgray}& \cellcolor{lightgray}{\textbf{77.3}} & \cellcolor{lightgray}{\textbf{75.2}} & \cellcolor{lightgray}{\textbf{89.7}} & \cellcolor{lightgray}& \cellcolor{lightgray}{\textbf{78.9}} & \cellcolor{lightgray}{\textbf{73.1}} & \cellcolor{lightgray}{\textbf{90.6}} & \cellcolor{lightgray}& \cellcolor{lightgray}{\textbf{74.5}} & \cellcolor{lightgray}{\textbf{72.1}} & \cellcolor{lightgray}{\textbf{89.2}} \\
\bottomrule
\end{tabular}
}
\vspace{-3pt}
\caption{Performance comparison on the gRefCOCO datasets using cIoU, gIoU, and N-Acc metrics. Models trained on our dataset (shaded in \smash{\colorbox{lightgray}{light gray}}) achieve the best performance, both with and without fine-tuning (ft) on gRefCOCO.}
\label{tab:grefcoco}
\end{table*}

\noindent\textbf{Human Annotations}  \ To ensure accurate model evaluation, we conduct rigorous human annotation on the test data. Each instruction pair includes an image, a question and one or multiple segmentation masks, each labeled with an object name. During annotation, annotators view the image, question, and all segmentation masks with object names simultaneously to ensure they have full contextual information. Annotators then examine each object within each segmentation mask, one by one, following the order of the mask index, to verify if it accurately and relevantly answers the question by selecting YES, NO or UNSURE. 
For hallucination mitigation, annotators assess three types of hallucinations for each image: 1) if the mentioned object itself is absent from the image, 2) if the relationship between objects is not present in the image, and 3) if the object's attribute incorrectly assigned or not present in the image. Each image-instruction pair is reviewed independently by two different human annotators, and only those verified as correct (YES) by both are included in the test set, 
resulting in the removal of 23.1\% of the original test data due to erroneous or unsure labels. We provide further details in the supplementary materials about the human annotation process.

%

\section{Experiments and Results}
\subsection{Experiment Setup}
\noindent\textbf{Implementation Details} To assess \dataset's effectiveness, we train two representative models: LISA~\citep{lai2024lisa} and PSALM~\citep{zhang2024psalm}. These models feature distinct architectures and backbones, highlighting the generality of our dataset. For a fair comparison, we keep all training hyperparameters and evaluation settings consistent with their original configurations, changing only the data by incorporating our own. 
\begin{itemize}
\item{\textbf{LISA} } \ LISA~\citep{lai2024lisa} builds on LLaVA~\citep{liu2023llava}, and incorporates a CLIP image encoder~\citep{radford2021learning} and Vicuna-7B~\citep{vicuna2023} as the LLM with a new segmentation token added to its vocabulary. During the forward pass, the last-layer embedding of the segmentation token is decoded into a segmentation mask using the fine-tuned SAM pixel decoder~\citep{kirillov2023sam}.
\item \textbf{PSALM} \ PSALM~\citep{zhang2024psalm} uses Phi-1.5 (1.5B)~\citep{li2023textbooks} as the LLM, a Swin Transformer~\citep{liu2021swin} as the vision encoder, and Mask2Former~\citep{cheng2022masked} as the pixel decoder, with 100 mask tokens. For referring expression tasks, the mask with the highest confidence score is selected as the prediction. For other tasks, all masks with confidence scores above 0.6 are chosen as the final predictions. 
\end{itemize}


\begin{table*}[t!]
\centering
\resizebox{0.975\linewidth}{!}{
\begin{tabular}{lp{0.05cm}llp{0.05cm}lp{0.05cm}lp{0.05cm}llp{0.05cm}ll}
\toprule
\multirow{2}{*}{\textbf{Method}} & & \multicolumn{2}{c}{\textbf{Multi-granular}} & & \textbf{Multi-object} & & \textbf{Hallucination} & & \multicolumn{2}{c}{\textbf{Reasoning}} & & \multicolumn{2}{c}{\textbf{Part Reference}} \\
\cmidrule{3-4} \cmidrule{6-6} \cmidrule{8-8} \cmidrule{10-11} \cmidrule{13-14}
  & & \textbf{Abstract} (gIoU) & \textbf{FG} (gIoU) & & \textbf{gIoU} & & \textbf{N-Acc} & & \textbf{gIoU} & \textbf{cIoU} & & \textbf{gIoU} & \textbf{cIoU} \\
\midrule
  GSVA & & 45.0 & 46.7  & & 30.4  & & \phantom{0}9.3  & & 40.1  & 41.2  & & 21.9 & 32.4 \\
  \rowcolor{lightgray}
GSVA-G5 & & 47.8 (\textcolor{blue}{\phantom{0}2.8$\uparrow$}) & 51.4 (\textcolor{blue}{\phantom{0}4.7$\uparrow$}) & & 39.6 (\textcolor{blue}{\phantom{0}9.2$\uparrow$}) & & 37.5 (\textcolor{blue}{28.2$\uparrow$}) & & 50.8 (\textcolor{blue}{10.7$\uparrow$}) & 48.9 (\textcolor{blue}{\phantom{0}7.7$\uparrow$}) & & 24.1 (\textcolor{blue}{2.2$\uparrow$}) & 35.6 (\textcolor{blue}{3.2$\uparrow$}) \\
  LISA & & 35.3 & 47.1 & & 33.1 & & \phantom{0}8.7 & & 42.2 & 49.9 & & 15.6 & 33.9 \\
\rowcolor{lightgray}
LISA-G5 & & 44.5 (\textcolor{blue}{\phantom{0}9.2$\uparrow$}) & 55.0 (\textcolor{blue}{\phantom{0}7.9$\uparrow$}) & & 39.2 (\textcolor{blue}{\phantom{0}6.1$\uparrow$}) & & 17.9 (\textcolor{blue}{\phantom{0}9.2$\uparrow$}) & & 45.1 (\textcolor{blue}{\phantom{0}2.9$\uparrow$}) & 55.8 (\textcolor{blue}{\phantom{0}5.9$\uparrow$}) & & 21.4 (\textcolor{blue}{5.8$\uparrow$}) & 35.2 (\textcolor{blue}{1.3$\uparrow$}) \\
PSALM & & 37.9 & 40.1 & & 27.3 & & 20.5 & & 53.7 & 53.9 & & 19.3 & 34.2 \\
\rowcolor{lightgray}
PSALM-G5 & & 58.7 (\textcolor{blue}{20.8$\uparrow$}) & 54.7 (\textcolor{blue}{14.6$\uparrow$}) & & 39.5 (\textcolor{blue}{12.2$\uparrow$}) & & 33.9 (\textcolor{blue}{13.4$\uparrow$}) & & 72.9 (\textcolor{blue}{19.2$\uparrow$}) & 72.2 (\textcolor{blue}{18.3$\uparrow$}) & & 27.4 (\textcolor{blue}{8.1$\uparrow$}) & 41.9 (\textcolor{blue}{7.6$\uparrow$}) \\
\bottomrule
\end{tabular}
}
\vspace{-3pt}
\caption{Performance comparison of models trained with and without proposed~\dataset, evaluated on the disjoint test set of~\dataset.}
\label{tab:ours_eval}
\end{table*}

\noindent\textbf{Evaluation Benchmarks and Metrics}  \ We evaluate on both public datasets and our more challenging Ground-V test set. Details for each benchmark are listed below:

\begin{itemize}

\item\textbf{(Generalized) referring expression segmentation}  \ We use the commonly adopted RefCOCO/+/g~\citep{yu2016refcoco, nagaraja2016modeling} to evaluate referring expression segmentation (RES) with cumulative IoU (cIoU). The UNC split is used for RefCOCO and RefCOCO+, while the UMD split is used for RefCOCOg. For generalized referring expression segmentation (GRES), we evaluate on gRefCOCO~\citep{liu2023gres}, which includes multi-target and no-target scenarios. The primary metrics are generalized IoU (gIoU), cIoU, and N-Accuracy (N-Acc) measuring the model's ability to identify no-target samples. For no-target samples, a prediction with no foreground pixels is considered a true positive (TP), while any prediction with foreground pixels is a false negative (FN). N-acc, defined as $\text{N-acc} = \frac{\text{TP}}{\text{TP} + \text{FN}} $, is often referred to as the True Positive Rate (TPR) in other domains, and it measures the model's ability to correctly identify no-target samples. For our benchmark, gIoU is the primary metric for segmentation accuracy, while N-Acc is used for the hallucination subset. Higher values for these metrics indicate better performance.

\item\textbf{Described object segmentation}  \
We evaluate the described object segmentation task using the Description Detection Dataset (D$^3$)~\citep{xie2023described}, reformulating it from detection to segmentation using the ground-truth segmentation masks. This task is more challenging than GRES, as it involves unrestricted language descriptions and expressions of absence (e.g., "man \textit{without} a hat"). We use gIoU, cIoU, and N-Acc as the evaluation metrics.


\item\textbf{\dataset \ (ours)} We divide the evaluation partition of the \dataset \ into five subsets: multi-granular, multi-object, hallucinated reference, reasoning, and part reference. The multi-granular subset, divided into abstract and fine-grained (FG) levels, is designed to evaluate the model’s ability to understand and segment concepts at different granularities and is assessed using the gIoU metric. The multi-object subset assesses segmentation of multiple objects, also using gIoU. The hallucinated reference subset measures the model's ability to avoid hallucinated segmentation with the N-Acc metric. The reasoning subset assesses reasoning skills through gIoU and cIoU. Finally, the part reference subset evaluates the model’s understanding of specific reference to object parts with both gIoU and cIoU metrics. This structured evaluation approach offers a thorough analysis of model performance across key challenges in visual grounding.

\end{itemize}

\noindent\textbf{Baselines} \
We compare both specialist models and VLM-based models on these benchmarks. For specialists, we report the performance of SEEM-L~\citep{zou2024seem}, PolyFormer-L~\citep{liu2023polyformer}, and UNINEXT-L/H~\citep{yan2023uninext} on RefCOCO/+/g. We also compare with representative SoTA VLMs such as LISA~\citep{lai2024lisa}, GLaMM~\citep{rasheed2024glamm}, u-LLaVA~\citep{xu2023u}, PerceptionGPT~\citep{pi2024perceptiongpt}, PixelLM~\citep{ren2024pixellm}, GSVA~\citep{xia2024gsva}, and PSALM~\citep{zhang2024psalm}. On the gRefCOCO dataset, we also include comparisons with specialist models trained specifically on this dataset, including MattNet~\citep{yu2018mattnet}, LTS~\citep{ding2021vision}, LAVT~\citep{yang2022lavt} and ReLA~\citep{liu2023gres}.

\subsection{Quantitative Results}
\noindent \textbf{Results on (G)RES benchmarks} \ 
The results on RefCOCO, RefCOCO+, and RefCOCOg are presented in Table~\ref{tab:refcoco}. Our model achieves state-of-the-art performance on most subsets, outperforming both specialist and VLM-based models. On gRefCOCO in Table~\ref{tab:grefcoco}, our model outperforms existing methods by a significant margin, both with and without post-hoc fine-tuning (ft), especially in N-Acc that measures the ability to abstain no-target instructions. This demonstrates that training on our dataset generalizes well to other benchmarks, as the diverse instructions, such as multi-granular text inputs, enhance the model's understanding of general referring expressions. Additionally, our hallucination mitigation data helps the model accurately identify when referenced objects are absent.

\noindent\textbf{Results on D$^3$ dataset} \
Table~\ref{tab:d3} presents the results on the D$^3$ dataset. After incorporating our proposed {\dataset} during training, both LISA and PSALM exhibit significant performance improvements across all metrics. Notably, PSALM trained on our dataset achieves 9.0\% and 23.5\% gains in gIoU and N-Acc, respectively, highlighting enhanced accuracy in understanding complex language descriptions and avoiding false segmentations of non-existent objects. These results demonstrate the efficacy of our constructed reasoning, multi-object, and hallucination data categories, which generalize well to unseen datasets.

\begin{table}[t!]
\centering
\resizebox{\linewidth}{!}{
\begin{tabular}{llll}
\toprule
\textbf{Model} & \textbf{gIoU} & \textbf{cIoU} & \textbf{N-Acc} \\
\midrule
LISA & 31.9 & 36.8 & 12.8 \\
\rowcolor{lightgray}
LISA-G5 & 35.3 (\textcolor{blue}{3.4$\uparrow$}) & 41.4 (\textcolor{blue}{4.6$\uparrow$}) & 26.5 (\textcolor{blue}{13.7$\uparrow$}) \\
PSALM & 49.5 & 46.0 & 42.5 \\
\rowcolor{lightgray}
PSALM-G5 & 58.5 (\textcolor{blue}{9.0$\uparrow$}) & 51.0 (\textcolor{blue}{5.0$\uparrow$}) & 66.0 (\textcolor{blue}{23.5$\uparrow$}) \\

\bottomrule
\end{tabular}
}
\vspace{-3pt}
\caption{Performance comparison on D$^{3}$ dataset with and without \dataset~during training. Models trained on our data (shaded in \smash{\colorbox{lightgray}{light gray}}) show significant improvement on this unseen dataset.}
\label{tab:d3}
\end{table}

\begin{table}[t!]
\centering
\begin{tabularx}{\columnwidth}{l*{2}{>{\centering\arraybackslash}X}}
\toprule
\textbf{Dataset} & \textbf{gIoU} & \textbf{cIoU} \\
\midrule
\dataset & 72.9 &  72.2  \\
Rephrased \dataset & 71.5  & 72.3  \\
\bottomrule
\end{tabularx}
\vspace{-3pt}
\caption{Performance comparison on \dataset~Reasoning subset before and after rephrasing, showing that our model is robust to language priors, with improvements not due to overfitting.}
\label{tab:rephrase}
\end{table}
\begin{table}[t!]
\centering
\resizebox{\linewidth}{!}{
\begin{tabular}{lllll}
\toprule
\textbf{Model} & \textbf{POPE} & \textbf{Vizwiz} & \textbf{MMBench} & \textbf{SciQA-Img}\\
\midrule
LLaVA-v1.5-7B & 85.9 &  50.0 &  64.3  & 66.8 \\
LLaVA-v1.5--7B-Ours & \textbf{86.2}  & \textbf{53.6} & \textbf{65.1} & \textbf{69.9}  \\
\bottomrule
\end{tabular}
}
\vspace{-3pt}
\caption{Performance comparison on general VQA tasks with and without \dataset~during LLaVA vision instruction tuning.} \label{tab:llava_vqa_bench}
\end{table}



\begin{figure*}[t!]
  \centering
  \includegraphics[width=1\textwidth]{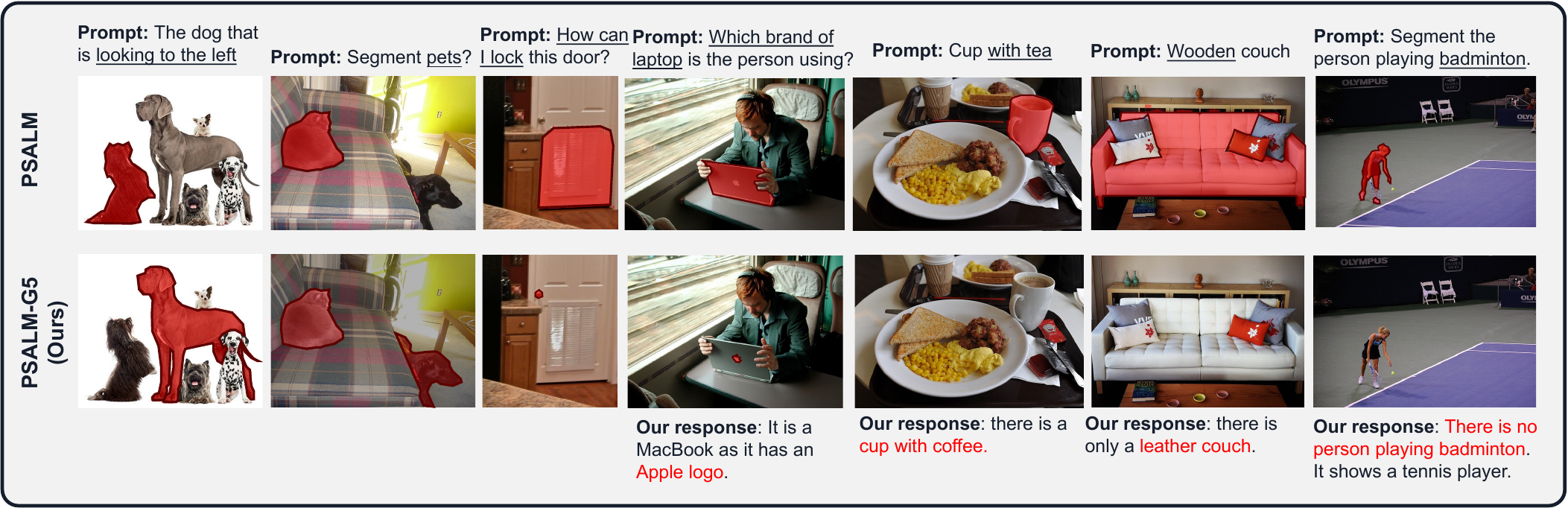}
  \caption{Our model demonstrates improved understanding of object attributes, enhanced reasoning capabilities, more precise segmentation of smaller object parts, and the ability to reject instructions when the target object is not present in the image. 
  }
  \label{fig:qualitative}
\end{figure*}

\noindent\textbf{Results on \dataset}
The results on the test partition of \dataset~are presented in Table~\ref{tab:ours_eval}. As shown, training on our dataset leads to significant improvements over the original PSALM across all subsets and metrics, with performance gains ranging from 10\% to 20\%. Specifically, our model shows notable enhancements in each of the five challenging dimensions considered in the curation of \dataset. These results underscore the value of our curated dataset in enhancing model performance, especially in handling complex visual scenarios and mitigating hallucinations.

\noindent\textbf{Generalization Beyond Language Priors} \ Since both the training and evaluation data in \dataset~are generated using the same workflow, we assess whether the performance improvements are due to overfitting to a consistent language style. To test this, we employ an off-the-shelf VLM, Qwen2-VL~\citep{wang2024qwen2vl}, to rephrase the instructions while ensuring that the underlying answers remain unchanged. We focus this experiment on the reasoning subset, where language input has a more significant impact on reasoning-based segmentation. For example, we rephrase the original instruction, ``\textit{What can we use to make a dinner? Find the objects},'' to ``\textit{Give a step-by-step guide for making a dinner. Segment the related objects in the image}.'' This allows us to evaluate if the model's performance is robust to variations in language style. As shown in Table~\ref{tab:rephrase}, the performance remains consistent after rephrasing, indicating that the model trained on our dataset is robust to variations in language priors. This suggests that the improvements are not merely due to overfitting to specific language styles but reflect the model’s ability to generalize to those challenging scenarios.

\noindent\textbf{Will generalist VLMs benefit from Ground-V?} We explore whether generalist VLMs like LLaVA~\citep{liu2023llava} can benefit from \dataset~by integrating its hallucination subset during the visual instruction tuning stage. This subset is text only, making it flexible for integration with existing visual instruction datasets, while the other subsets include text and segmentation mask pairs\footnote{Although the other subsets could also be modified to text-only and incorprated into LLaVA training, this is outside the scope of this work.}. We evaluate on public VQA benchmarks such as POPE~\citep{li2023evaluating}, Vizwiz~\citep{gurari2018vizwiz}, MMBench~\citep{liu2024mmbench} and SciQA~\citep{lu2022learn} with results compiled in Table~\ref{tab:llava_vqa_bench}. Remarkably, we observe consistent performance improvements across all benchmarks, demonstrating that even generalist VLMs can benefit from \dataset.


\subsection{Qualitative Results}
Figure~\ref{fig:qualitative} shows qualitative comparisons between PSALM (top row) and our model (bottom row) under the same instructions. Compared to the baseline PSALM, our model effectively navigates multi-granularity, multi-object contexts, and part-based references, accurately avoiding incorrect segmentation of irrelevant objects under ambiguous instructions. For example, in the first and second columns, our model accurately interprets instructions at different specificity levels: it correctly segments the left-facing dog in the first column (fine-grained instance-level reference) and includes both the cat and dog in the second column under the broader instruction for ``pets" (abstract category). The third to fourth columns demonstrate our model’s reasoning abilities, such as in the fourth column, where it accurately identifies and segments part of a laptop as a MacBook based on the Apple logo. In the fifth to seventh columns, our model effectively mitigates hallucination by rejecting an misleading instruction, noting the absence of mentioned objects (e.g., ``person playing badminton'') in the image, while PSALM mistakenly segments conceptually similar objects (e.g., ``person playing tennis''). 

\section{Discussion}
In this paper, we explore a data-centric approach to enhance VLMs' ability to ground complex instructions with nuanced object references by developing an instruction-following data generation workflow and introducing a new dataset called \dataset. By systematically addressing five critical challenges in open-world visual grounding -- multi-object interactions, multi-granularity, hallucination prevention, part-level reference, and reasoning -- \dataset~significantly improves the segmentation accuracy of VLMs across challenging scenarios. Experiment results demonstrated that models trained on \dataset~achieve significant performance gains, with the best-performing models surpassing previous state-of-the-art results on standard benchmarks such as RefCOCO/g/+, gRefCOCO, and $D^3$. These findings highlight {\dataset}'s potential to bridge the gap between linguistics and pixel-level visual perception, advancing visual grounding in complex real-world applications. We hope our work encourages more scalable approaches to augment existing abundant computer vision data to unlock new capabilities in multimodal grounding.

\keypoint{Limitations and Future Work} While models trained on our dataset show strong performance,  we acknowledge the presence of noise in the training data due to its automatic generation by off-the-shelf models. Future work could explore ways to mitigate data noise in an automated manner. 

\clearpage

{
    \small
    \bibliographystyle{ieeenat_fullname}
    \bibliography{main}
}

\setcounter{page}{1}
\appendix
\maketitlesupplementary

\section{Implementation Details}

\subsection{Datasets}
For a fair comparison, we adopt datasets originally used in LISA~\citep{lai2024lisa} and PSALM~\citep{zhang2024psalm} for training. They are listed in Table~\ref{tab:dataset_lisa} and~\ref{tab:dataset_psalm} respectively. 

\begin{table}[h]
\centering
\resizebox{\linewidth}{!}{%
\begin{tabular}{lp{0.7\linewidth}}
\toprule
\textbf{Task} & \textbf{Datasets} \\
\midrule
Semantic Segmentation & 
ADE20K~\citep{zhou2019ade20k}, COCO-Stuff~\citep{caesar2018cocostuff}, PACO-LVIS~\citep{ramanathan2023paco},
PartImageNet~\citep{he2022partimagenet} \\
Referring Segmentation & 
refCOCO/+/g~\citep{yu2016refcoco, mao2016generation}, refCLEF~\citep{liao2020real} \\
VQA & 
LLaVA Instruct-150k~\citep{liu2023llava} \\
Reasoning Segmentation & 
ReasonSeg~\citep{lai2024lisa} \\
\bottomrule
\end{tabular}%
}
\caption{Training datasets of LISA.}
\label{tab:dataset_lisa}
\end{table}

\begin{table}[h]
\centering
\begin{tabular}{ll}
\toprule
\textbf{Task} & \textbf{Datasets} \\
\midrule
Generic Segmentation & 
COCO-Panoptic~\citep{lin2014mscoco} \\
Referring Segmentation & 
refCOCO/+/g~\citep{yu2016refcoco, mao2016generation}\\
VQA & 
LLaVA-v1.5-665k~\citep{liu2024improved} \\
\bottomrule
\end{tabular}%
\caption{Training datasets of PSALM.}
\label{tab:dataset_psalm}
\end{table}

\subsection{Training Configurations}
We follow the official implementation and training configurations of LISA\footnote{https://github.com/dvlab-research/LISA} and PSALM\footnote{https://github.com/zamling/PSALM} for training. The hyper-parameters of LISA and PSALM training are listed in Table~\ref{tab:hp_lisa} and~\ref{tab:hp_psalm} respectively. For ablation experiments in Section~\ref{sec:ablation}, we train models for 3 epochs and keep the other settings the same. All models are trained on a single 8$\times$A100 40GB machine.

\begin{table}[h]
\centering
\begin{tabular}{lr}
\toprule
\textbf{Parameters} & \textbf{Value} \\
\midrule
Optimizer & AdamW~\citep{loshchilov2019adamw} \\
Learning Rate & $3 \times 10^{-4}$ \\
Batch Size Per GPU & 10 \\
Gradient Accumulation Steps & 2 \\
Number of Epochs & 10 \\
Learning Rate Schedule & WarmupDecayLR \\
Weight Decay & 0.0 \\
Warmup Ratio & 0.03 \\
LoRA $\alpha$ & 64 \\
Image Size & $224 \times 224$ \\
\bottomrule
\end{tabular}
\caption{Training hyper-parameters for LISA.}
\label{tab:hp_lisa}
\end{table}

\begin{table}[h]
\centering
\begin{tabular}{lr}
\toprule
\textbf{Parameters} & \textbf{Value} \\
\midrule
Optimizer & AdamW~\citep{loshchilov2019adamw} \\
Learning Rate & $4 \times 10^{-5}$ \\
Batch Size Per GPU & 8 \\
Gradient Accumulation Steps & 1 \\
Number of Epochs & 10 \\
Learning Rate Schedule & Cosine Decay \\
Weight Decay & 0.0 \\
Warmup Ratio & 0.03 \\
$\beta_1$ & 0.9 \\
$\beta_2$ & 0.999 \\
Image Size & $1024 \times 1024$ \\
\bottomrule
\end{tabular}
\caption{Training hyper-parameters for PSALM.}
\label{tab:hp_psalm}
\end{table}

\section{Experiments}
\subsection{Evaluation on ReasonSeg and MUSE}
We further evaluate the reasoning-based segmentation benchmarks ReasonSeg~\citep{lai2024lisa} and MUSE~\citep{ren2024pixellm} using LISA and LISA-G5 trained on \dataset. The test set results are presented in Table~\ref{tab:reasonseg} and Table~\ref{tab:muse}. Models trained on \dataset achieve a 1.8\% gIoU improvement on ReasonSeg and a 4.9\% gIoU improvement on MUSE. Notably, the evaluation on MUSE is conducted in a zero-shot setting, and the observed improvement highlights the effectiveness of training on \dataset, particularly the multi-object and reasoning subsets, as MUSE involves scenarios with multiple objects.

\begin{table}[t!]
\centering
\begin{tabular}{lll}
\toprule
\textbf{Models} & \textbf{gIoU} & \textbf{cIoU} \\
\midrule
LISA &  50.3 & 49.8  \\
LISA-G5 &  52.1 (\textcolor{blue}{1.8$\uparrow$}) & 51.9 (\textcolor{blue}{2.1$\uparrow$}) \\
\bottomrule
\end{tabular}
\vspace{-3pt}
\caption{Performance comparison on ReasonSeg test set.}
\label{tab:reasonseg}
\end{table}

\begin{table}[t!]
\centering
\begin{tabular}{lll}
\toprule
\textbf{Models} & \textbf{gIoU} & \textbf{cIoU} \\
\midrule
LISA &  12.6 &  27.2\\
LISA-G5 &  17.5 (\textcolor{blue}{4.9$\uparrow$}) &  30.4 (\textcolor{blue}{3.2$\uparrow$}) \\
\bottomrule
\end{tabular}
\vspace{-3pt}
\caption{Performance comparison on MUSE test set (zero-shot).}
\label{tab:muse}
\end{table}

\subsection{Ablations on Data Components}
\label{sec:ablation}
We conduct two ablation studies to further understand the introduced \dataset. We conduct experiments with the PSALM model. Note that we only train 3 epochs for ablation studies and thus the performance is not comparable with the performance in the main text.

\begin{figure}[t]
  \centering
  \includegraphics[width=\columnwidth]{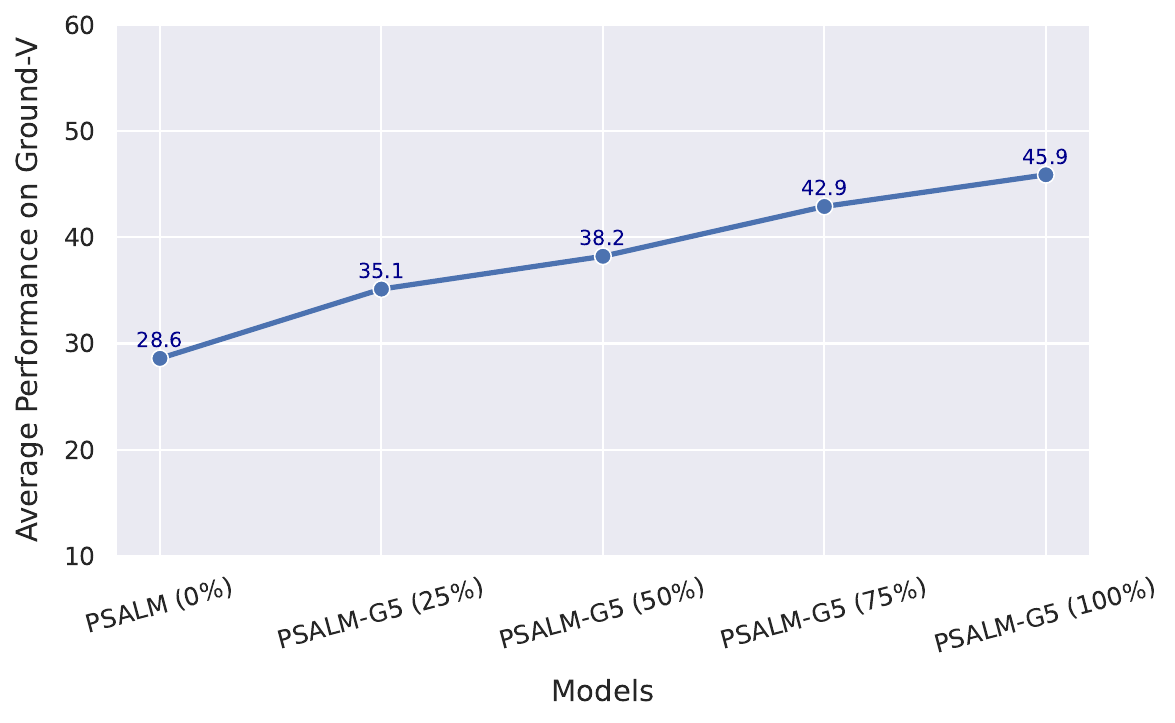}
  \caption{Performance of models trained with different proportions of \dataset. The average performance is calculated by averaging the gIoU of all subsets.
  }
  \label{fig:scale_eval}
\end{figure}

\keypoint{Scaling of the data samples} 
We first examines the impact of dataset scaling on model performance. Models are trained on subsets comprising 25\%, 50\%, 75\%, and 100\% of the data to assess the effectiveness of varying data volumes. The trained models are assessed on the \dataset~test set, with the average performance across different subsets presented in Figure\ref{fig:scale_eval}, and detailed results provided in Table~\ref{tab:scaling_eval}. As shown in Figure~\ref{fig:scale_eval}, increasing the amount of data used during training consistently improves evaluation performance, highlighting the effectiveness of our dataset.

\begin{table*}[t!]
\centering
\resizebox{\linewidth}{!}{
\begin{tabular}{lp{0.05cm}ccp{0.05cm}cp{0.05cm}cp{0.05cm}ccp{0.05cm}cc}
\toprule
\textbf{Method} & & \multicolumn{2}{c}{\textbf{Multi-granular}} & & \textbf{Multi-object} & & \textbf{Hallucination} & & \multicolumn{2}{c}{\textbf{Reasoning}} & & \multicolumn{2}{c}{\textbf{Part Reference}} \\
\cmidrule{3-4} \cmidrule{6-6} \cmidrule{8-8} \cmidrule{10-11} \cmidrule{13-14}
  & & \textbf{Abstract} (gIoU) & \textbf{Fine-grained} (gIoU) & & \textbf{gIoU} & & \textbf{N-Acc} & & \textbf{gIoU} & \textbf{cIoU} & & \textbf{gIoU} & \textbf{cIoU} \\
\midrule
PSALM (0\%) & & 27.3 & 29.9 & & 17.5 & & 14.7 & & 43.5 & 41.1 & & 12.3  & 24.9 \\
PSALM-G5 (25\%) & & 34.1 & 37.3 & & 23.7 & & 16.9 & & 50.2 & 50.8 & & 14.6 & 28.7 \\
PSALM-G5 (50\%) & & 38.6 & 39.5 & & 25.8 & & 19.7 & & 54.3 & 53.9 & & 17.1 & 29.4 \\
PSALM-G5 (75\%) & & 45.9 & 43.6 & & 28.9 & & 22.5 & & 60.3 & 59.8 & & 18.8 & 31.2 \\
\rowcolor{lightgray}
PSALM-G5 (100\%) & & 47.8  & 51.9  & & 30.5  & & 23.3  & & 62.2  & 61.9  & & 20.9  & 33.1  \\

\bottomrule
\end{tabular}
}
\vspace{-3pt}
\caption{Performance comparison of models trained with different proportions of \dataset, evaluated on the disjoint test set of~\dataset.}
\label{tab:scaling_eval}
\end{table*}

\begin{figure}[t]
  \centering
  \includegraphics[width=\columnwidth]{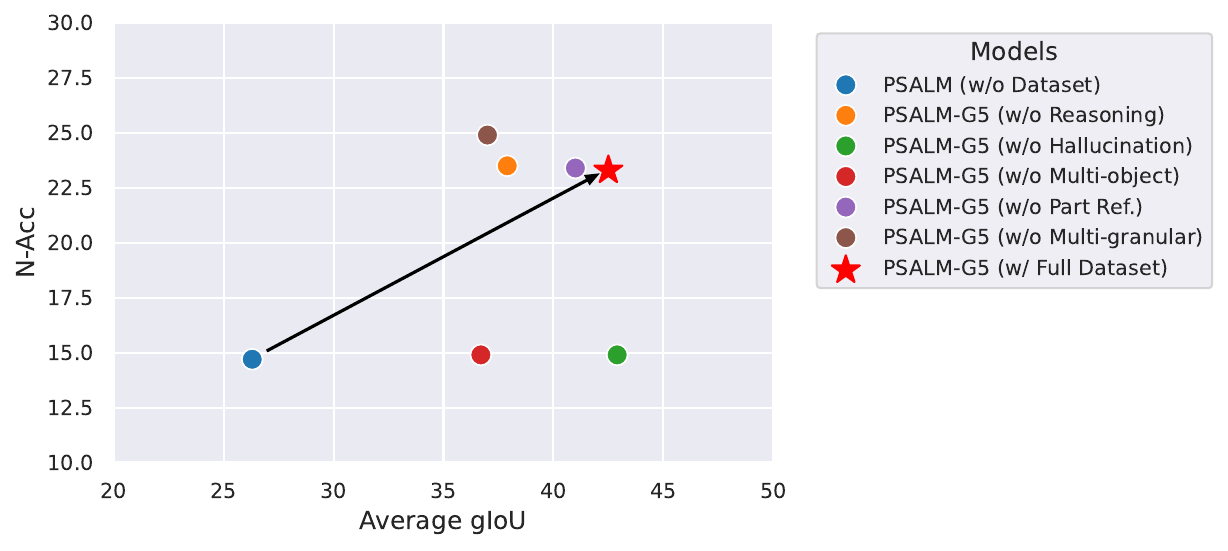}
  \caption{Performance of models trained on different subsets. X-axis measures how accurately the models can segment the objects and Y-axis measures how well the models can abstain the non-existing objects. The model trained on full \dataset~is on the Pareto curve and obtains significantly improved abilities to handle both scenarios.
  }
  \label{fig:subset_contrib}
\end{figure}

\keypoint{Contribution of different subsets} 
Next we explore the importance of different subsets within \dataset. This is achieved by training five models, each with one subset removed, alongside models trained with and without the full \dataset, to assess their contributions to overall performance. As shown in Figure~\ref{fig:subset_contrib}, the model trained on the full \dataset~lies on the Pareto curve, demonstrating significantly improved capabilities in both accurate segmentation and effective abstention from non-existent objects. 

While decreased performance on the corresponding test set when a subset is removed from training is expected, we observe several interesting patterns: (1) Most subsets are mutually beneficial: For example, removing the multi-object subset also degrades performance on the reasoning subset, as the reasoning subset often involves multiple objects, and training on the multi-object subset enhances the model’s ability to handle such scenarios—and vice versa. (2) Some subsets can introduce trade-offs: For instance, the performance on the fine-grained subset decreases when hallucination examples are included during training. This suggests that the hallucination subset can occasionally make the model overly cautious, leading to the rejection of existing objects. However, despite this trade-off, including hallucination examples in training remains valuable, as it significantly improves overall performance and helps mitigate hallucination-related errors. The detailed results are shown in Table~\ref{tab:ablation_subset}.

\begin{table*}[t!]
\centering
\resizebox{\linewidth}{!}{
\begin{tabular}{lp{0.05cm}ccp{0.05cm}cp{0.05cm}cp{0.05cm}ccp{0.05cm}cc}
\toprule
\textbf{Method} & & \multicolumn{2}{c}{\textbf{Multi-granular}} & & \textbf{Multi-object} & & \textbf{Hallucination} & & \multicolumn{2}{c}{\textbf{Reasoning}} & & \multicolumn{2}{c}{\textbf{Part Reference}} \\
\cmidrule{3-4} \cmidrule{6-6} \cmidrule{8-8} \cmidrule{10-11} \cmidrule{13-14}
  & & \textbf{Abstract} (gIoU) & \textbf{Fine-grained} (gIoU) & & \textbf{gIoU} & & \textbf{N-Acc} & & \textbf{gIoU} & \textbf{cIoU} & & \textbf{gIoU} & \textbf{cIoU} \\
\midrule
PSALM (w/o \dataset) & & 27.3 & 29.9 & & 17.5 & & 14.7 & & 43.5 & 41.1 & & 12.3 & 24.9 \\
PSALM-G5 (w/o Reasoning) & & 46.5 & 50.0 & & 25.4 & & 23.5 & & 44.1 & 41.9 & & 20.1 & 33.0 \\
PSALM-G5 (w/o Hallucination) & & 46.7 & \textbf{52.6} & & 29.4 & & 14.9 & & 60.1 & 61.0 & & \textbf{21.4} & \textbf{34.5} \\
PSALM-G5 (w/o Multi-object) & & 40.4 & 48.3 & & 18.2 & & 14.9 & & 52.1 & 55.3 & & 19.6 & 33.0 \\
PSALM-G5 (w/o Part Ref.) & & \textbf{48.0} & 52.1 & & 29.8 & & 23.4 & & 61.9 & 61.4 & & 12.5 & 25.2 \\
PSALM-G5 (w/o Multi-granular) & & 34.6 & 36.7 & & 27.3 & & \textbf{24.9} & & 62.0 & 61.7 & & 20.8 & 32.4 \\
\rowcolor{lightgray}
PSALM-G5 (w/ full \dataset) & & 47.8 & 51.9 & & \textbf{30.5} & & 23.3 & & \textbf{62.2} & \textbf{61.9} & & 20.9 & 33.1 \\
\bottomrule
\end{tabular}
}
\vspace{-3pt}
\caption{Performance comparison of models trained with different subsets of \dataset, evaluated on the disjoint test set of~\dataset. We remove one subset at a time for training.}
\label{tab:ablation_subset}
\end{table*}

\section{Dataset Details}
\subsection{Statistics}
\begin{figure*}[ht!]
  \centering
  \includegraphics[width=\textwidth]{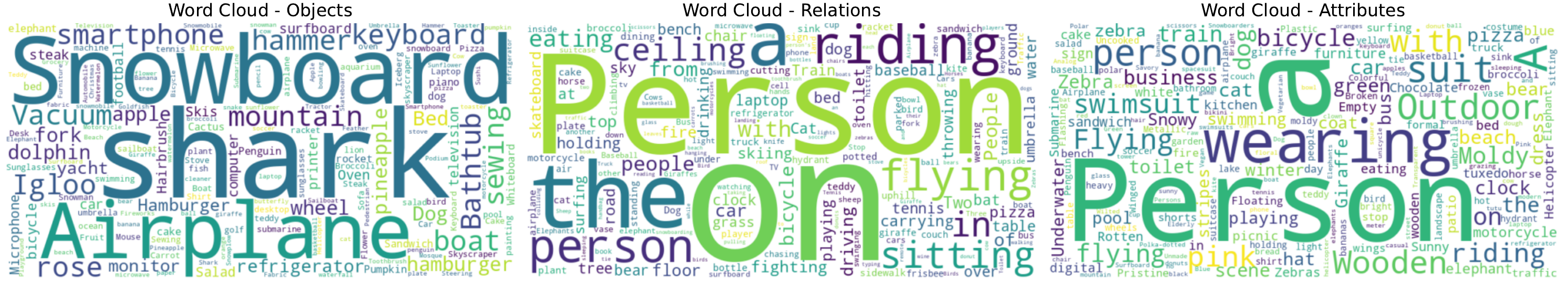}
  \includegraphics[width=\textwidth]{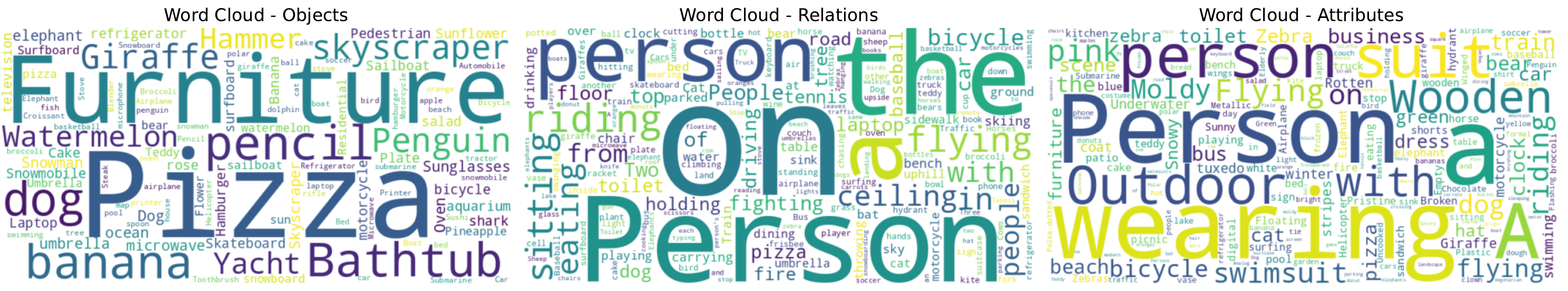}
  
  \caption{Word Cloud visualizations for training (top) and evaluation (bottom) sets of hallucination mitigation subsets. We show word clouds for object, relation, and attribute hallucination.}
  \vspace{-5pt}
  \label{fig:word_hallucination}
\end{figure*}

\begin{figure*}[ht!]
  \centering
  \includegraphics[width=\textwidth]{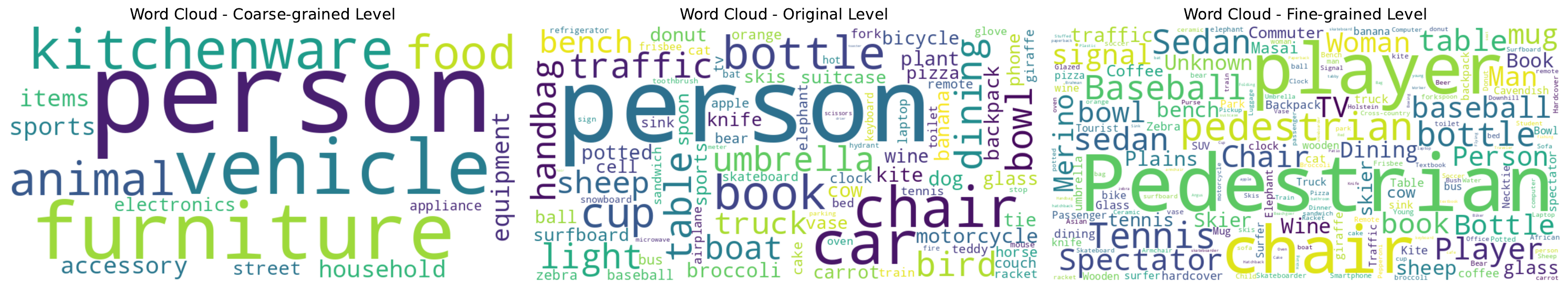}
  
  \includegraphics[width=\textwidth]{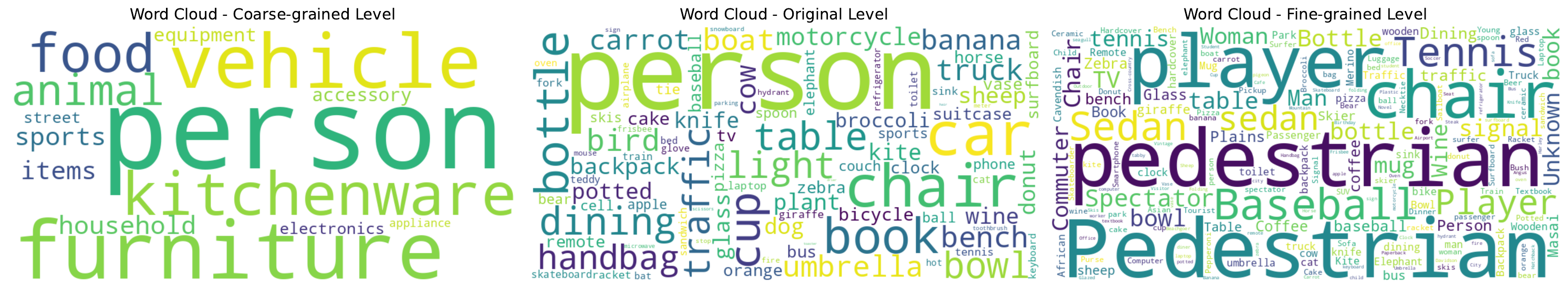}
  
  \caption{Word Cloud visualizations for training (top) and evaluation (bottom) sets of multi-granular subsets. We show word clouds for coarse-grained, original, fine-grained categories.}
  \vspace{-5pt}
  \label{fig:word_multigranular}
\end{figure*}

\begin{figure*}[t!]
  \centering
  \includegraphics[width=\textwidth]{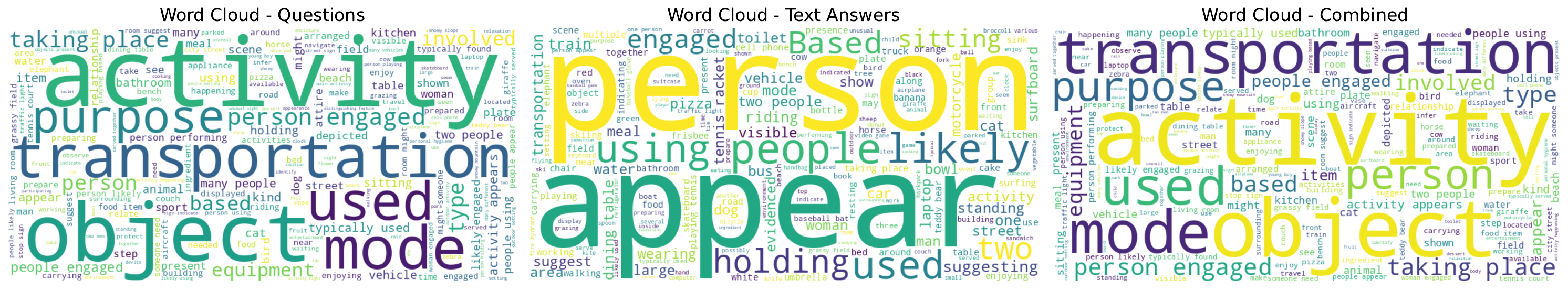}
  \includegraphics[width=\textwidth]{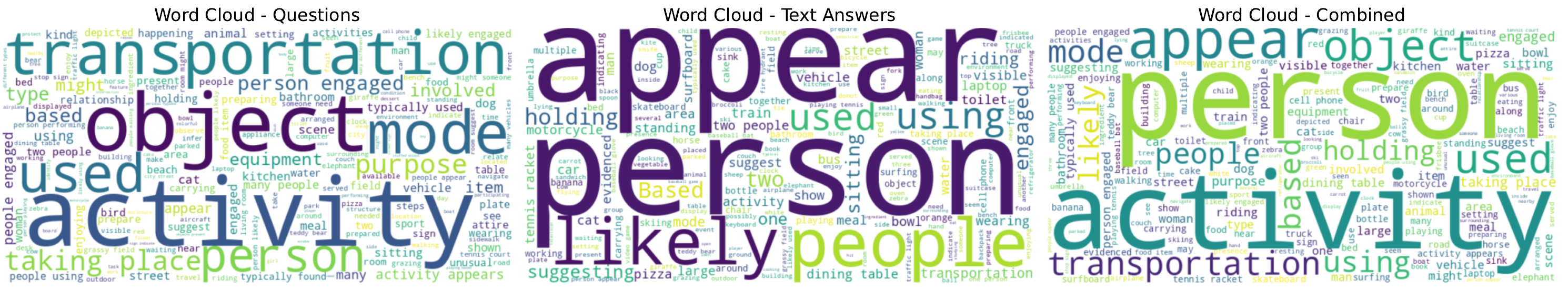}
  
  \caption{Word Cloud visualizations for training (top) and evaluation (bottom) sets of reasoning subsets. We show word clouds for questions, answers, as well as the combination of both.}
  \vspace{-5pt}
  \label{fig:word_reasoning}
\end{figure*}

\begin{figure*}[t!]
  \centering
  \includegraphics[width=\textwidth]{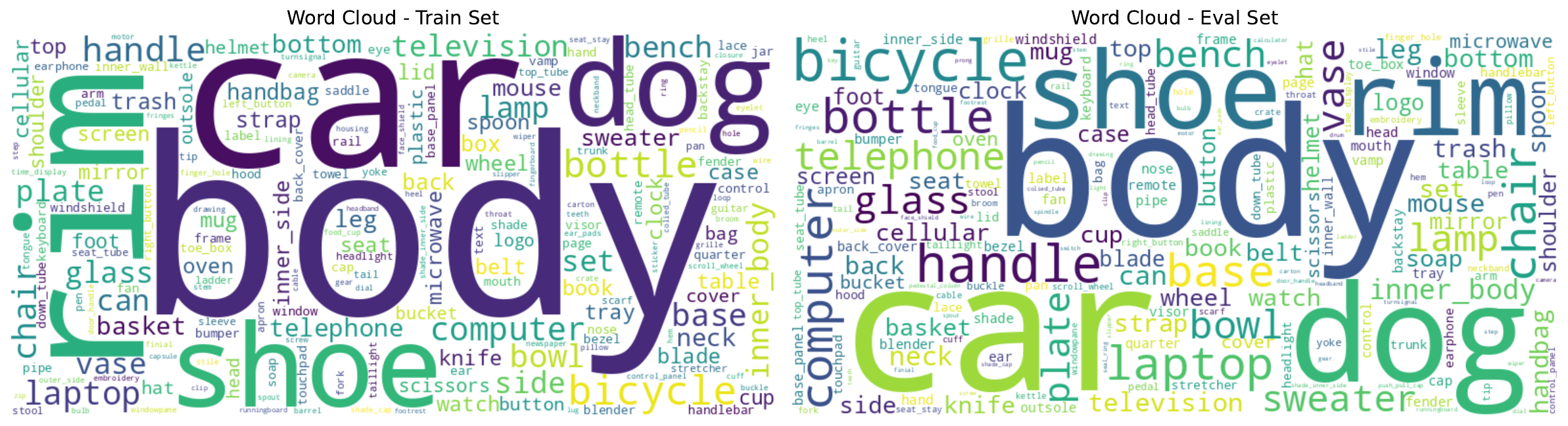}
  \caption{Word Cloud visualizations for training (left) and evaluation (right) sets of part reference subsets. }
  \label{fig:word_part}
\end{figure*}

\begin{figure*}[t!]
  \centering
  \includegraphics[width=\textwidth]{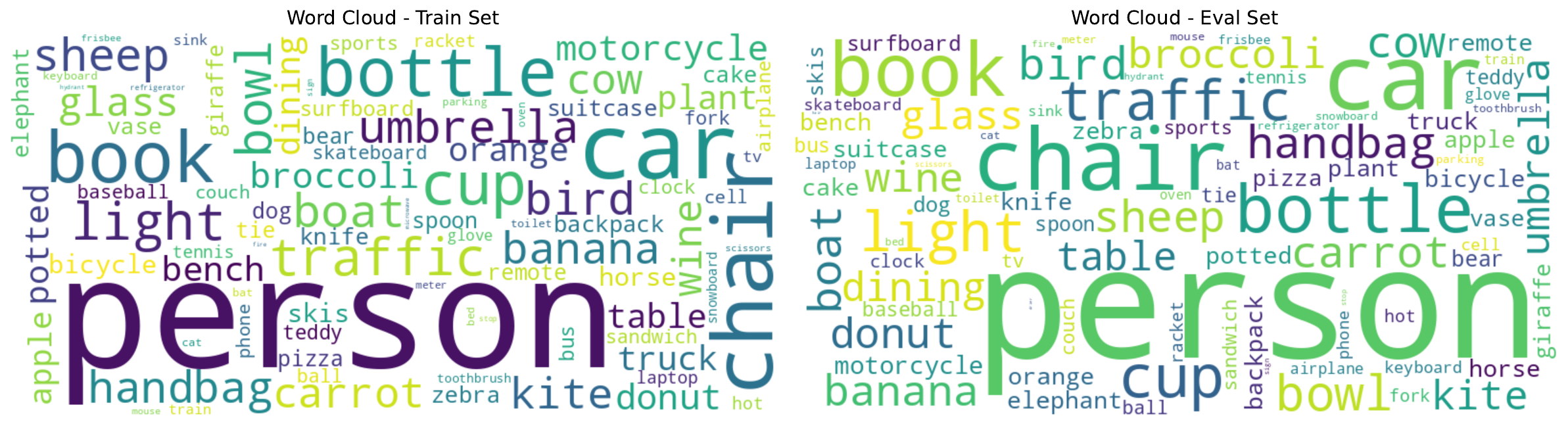}
  \caption{Word Cloud visualizations for training (left) and evaluation (right) sets of multi-object subsets. }
  \label{fig:word_multiobject}
\end{figure*}

We visualize the word frequencies of all subsets of \dataset~in Figure~\ref{fig:word_hallucination} to~\ref{fig:word_multiobject}. As it can be seen from these figures, the word distributions of our dataset are highly diverse across subsets.

\subsection{Human Annotation Process}
The human annotation process was meticulously designed to ensure the creation of high-quality, consistent class labels and accurate segmentation mask annotations for the test partition of \dataset. To minimize bias, the annotation was carried out by 20 external annotators who were independent of the authors. Prior to the full annotation phase, all annotators participated in training sessions and pilot trials to familiarize themselves with the task. Continuous quality control measures, including spot-checks and feedback loops, were implemented to address corner cases and enhance annotators' understanding of the task.

\begin{figure*}[ht!]
  \centering
  \includegraphics[width=\textwidth]{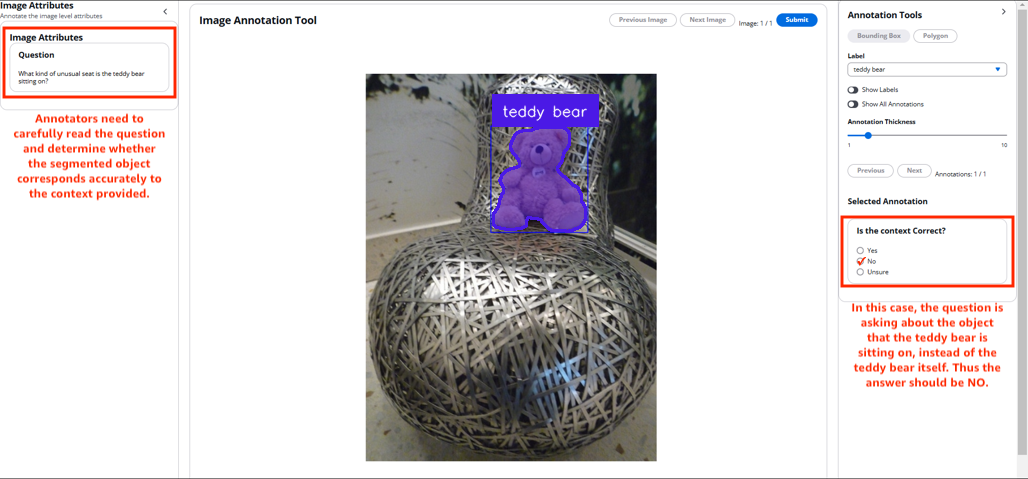}
  \caption{Human annotation interface for \dataset test data that resemble reasoning segmentation. In this case, the question is asking about the object that the teddy bear is sitting on instead of the teddy bear itself. Thus the answer should be NO. }
  \label{fig:anno_interface_bear}
\end{figure*}

\begin{figure*}[ht!]
  \centering
  \includegraphics[width=\textwidth]{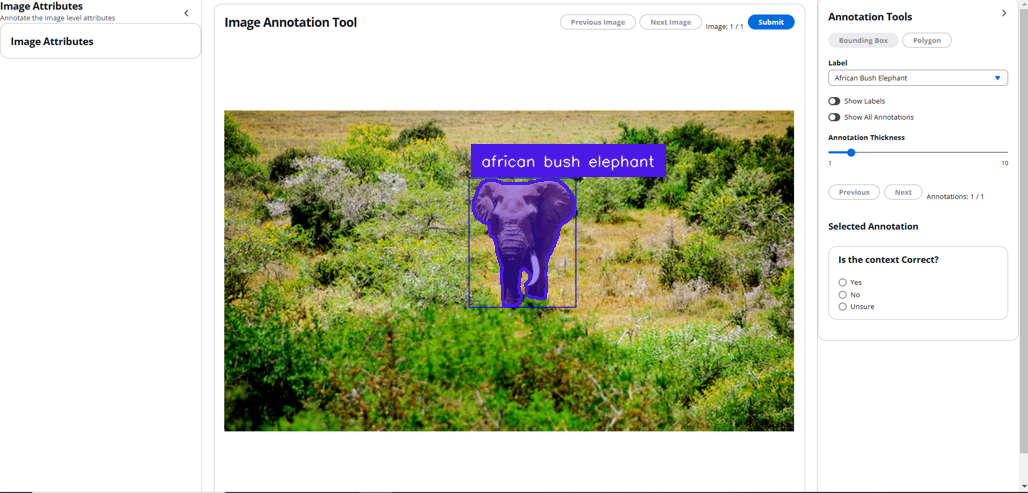}
  \caption{Human annotation interface for \dataset test data that resemble multi-granularity referring. In this example, the label ``African Bush Elephant'' is more specific than the general label ``elephant'' that most people would typically use. The data annotator needs to evaluate whether this finer-grained label accurately matches the segmented object.}
  \label{fig:anno_interface_elephant}
\end{figure*}

\begin{figure*}[ht!]
  \centering
  \includegraphics[width=\textwidth]{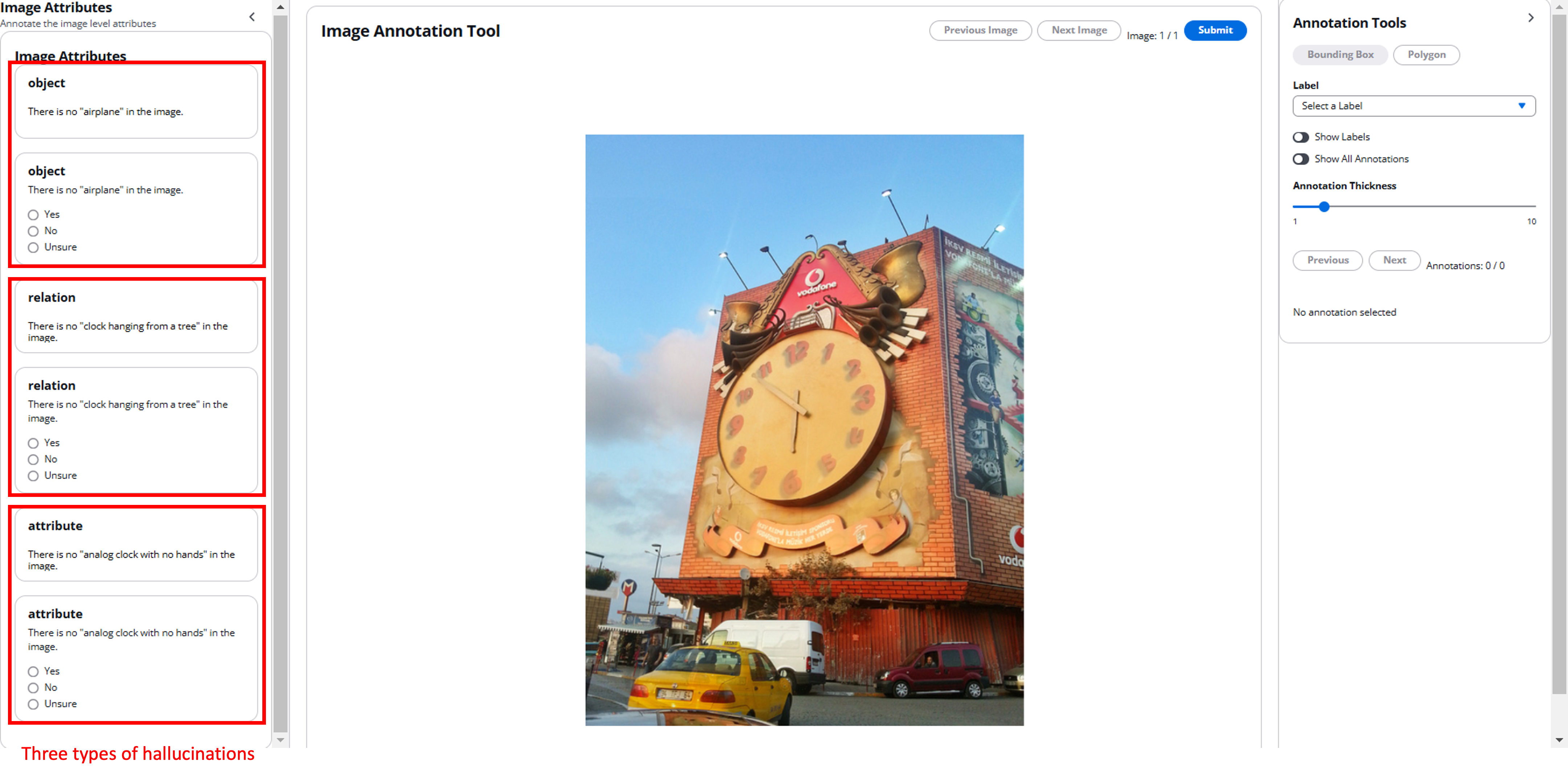}
  \caption{Human annotation interface for \dataset test data that resemble complex referrings that that could lead to hallucinations in segmentation. In this example, the annotator is asked to assess three types of hallucinations for each image: 1) if the mentioned object itself is absent from the image, 2) if the relationship between objects is not present in the image, and 3) if the object’s attribute incorrectly assigned or not present in the image. Each hallucination question is answered individually. }
  \label{fig:anno_interface_hallucination}
\end{figure*}

The annotation interface, illustrated in \cref{fig:anno_interface_bear,fig:anno_interface_elephant,fig:anno_interface_hallucination}, was specifically designed to support human annotations. Each annotation instance included an image, a question, and one or more segmentation masks, each associated with an object name. Annotators were presented with the image, question, and all segmentation masks along with their corresponding object names simultaneously, providing complete contextual information. Annotators examined each segmentation mask individually, following the order of the mask index, and determined whether the object accurately and relevantly answered the question by selecting one of three options: YES, NO, or UNSURE. They adhered to comprehensive guidelines, which covered definitions of the five challenges, examples for clarification, and detailed instructions for resolving ambiguous cases. If annotators were unsure about the correctness of a label (e.g., unfamiliar with a "Maine Coon cat"), they were instructed to select UNSURE by default. Additionally, if any polygon node in a segmentation mask deviated by more than 5\% of the object’s size from its actual boundaries, they selected NO for segmentation verification. In practice, the 5\% rule was applied based on the annotators' judgment and visual assessment. However, all annotators underwent training to ensure their evaluations aligned consistently with the requirements.

Each image-question pair was independently reviewed by two annotators. To maintain consistency and validate the process, periodic random audits were conducted by a third annotator, with different annotators auditing different instances. These audits revealed that when both initial annotators selected YES, there was a 95\%+ probability that the third annotator would also select YES. Based on this finding and budget constraints, we used two annotators for each annotation instance. Only instances where both annotators agreed on YES were included in the final test set, ensuring high confidence in the data's correctness. As a result, 23.1\% of the original test data was excluded due to erroneous or UNSURE labels. The remaining data is used as the test partition of \dataset.

\section{Prompt Details}

\onecolumn
\label{app:exp}
\subsection{Prompts for Data Curation}
We present the detailed prompts used for data curation below.

\keypoint{Prompt for attribute-level hallucination generation}

\begin{mdframed}[backgroundcolor=gray!20]
\setlength{\parindent}{0pt}
\textbf{System: } You are a helpful assistant who can help with analyze image. 
    You will be provided with an image containing certain objects and a set of object classes.
    You need to analyze the attributes of the objects and generate an exiting object (from the given classes) with different attribute that is not present in the image.
    \newline\newline
    Few-shot Examples:
    \newline\newline
    **Image 1**\newline
    - Present objects: Lemon.\newline
    - Attribute (you need to analyze the image): Yellow lemon.\newline
    - Negative sample: Green lemon.
    \newline\newline
    **Image 2**\newline
    - Present objects: person\newline
    - Attribute (you need to analyze the image): Person wearing a blue shirt.\newline
    - Negative sample: Person wearing a red shirt.\newline
    Note: in this case, you need to ensure there is no person wearing red shirt in the image.
    \newline\newline
    **Image 3**\newline
    - Present objects: car, road, person\newline
    - Attribute (you need to analyze the image): A shiny silver car parked in a driveway.\newline
    - Negative sample: A matte black car.
    \newline\newline
\textbf{User}: Now for the following image, generate an attribute-level negative sample. Only provide a attribute that is not present in the image, which would be clearly incorrect to request for segmentation. Do not generate reasoning process.
    Generate a Json dictionary in the format of {{"attribute": "the generated negative class"}}. Make sure the generated negative sample is realistic. If you think you cannot think of a good example in terms of attribute or it is unrealistic, output {{"attribute": "Unknown"}}.
    Generate Json only, strictly no other texts. Be concise in the generated class.
    \newline\newline
    **Query Image**  \newline
    - Present objects: [objects]

\textbf{Assistant}: 

\end{mdframed}

\newpage
\keypoint{Prompt for relation-level hallucination generation}

\begin{mdframed}[backgroundcolor=gray!20]
\setlength{\parindent}{0pt}
\textbf{System: } You are a helpful assistant who can help with analyze image. 
    You will be provided with an image containing certain objects and a set of object classes. Note that the object classes may not be comprehensive and you should also pay attention to the image.
    You need to analyze the relations between objects and generate a relation that is not present in the image with existing object(s). It can also be spatial relation.
    Generate the object first, and then generate the relation, i.e., we should clearly know the main object from the generated phase.
\newline\newline
    Few-shot Examples:
\newline\newline
    **Image 1**\newline
    - Present objects: child, horse\newline
    - Relation (you need to analyze the image): A child sitting beside a sleeping horse.\newline
    - Negative sample: Child who is riding the horse.\newline
\newline\newline
    **Image 2**\newline
    - Present objects: person, person\newline
    - Relation (you need to analyze the image): Two people walking in opposite directions\newline
    - Negative sample: Two people shaking hands.
\newline\newline
    **Image 3**\newline
    - Present objects: dog\newline
    - Relation (you need to analyze the image): dog on the left.\newline
    - Negative sample: dog on the right.\newline
    Note: in this case, you need to ensure there is no dog on the right in the image.
    \newline\newline
\textbf{User}: Now for the following image, generate an relation-level negative sample. Only provide a relation that is not present in the image, which would be clearly incorrect to request for segmentation. Do not generate reasoning process.
    Generate a Json dictionary in the format of {{"relation": "the generated negative class"}}. Make sure the generated negative sample is realistic. If you think you cannot think of a good example in terms of relation or it is unrealistic, output {{"relation": "Unknown"}}.
    Generate Json only, strictly no other texts. Be concise in the generated class.\newline
    **Query Image**  \newline
    - Present objects: [objects]

\textbf{Assistant}: 
\end{mdframed}

\newpage
\keypoint{Prompt for object-level hallucination generation}

\begin{mdframed}[backgroundcolor=gray!20]
\setlength{\parindent}{0pt}
\textbf{System: } You are a helpful assistant who can help with analyze image. 
    You will be provided with an image containing certain objects and a set of object classes. Note that the object classes may not be comprehensive and you should also pay attention to the image.
    Your task is to generate a negative class that is not present in the image. 
    \newline\newline
    Few-shot Examples:
\newline\newline
    **Image 1**\newline
    - Present objects: Airplane, Cloud, Runway\newline
    - Negative sample: Laptop
    \newline\newline
    **Image 2**\newline
    - Present objects: Chair, Laptop, Cup, Plate\newline
    - Negative sample: Television
\newline\newline
    **Image 3**\newline
    - Present objects: boots, businessman\newline
    - Negative sample: basketball shoes, businesswoman
\newline\newline
\textbf{User}: Now for the following image, generate an relation-level negative sample. Only provide a object that is not present in the image, which would be clearly incorrect to request for segmentation. Do not generate reasoning process.
    Generate a Json dictionary in the format of {{"object": "the generated negative class"}}. Make sure the generated negative sample is realistic. If you think you cannot think of a good example in terms of object or it is unrealistic, output {{"object": "Unknown"}}.
    Generate Json only, strictly no other texts. Be concise in the generated class.
    **Query Image**  \newline
    - Present objects: [objects]
    \newline
\textbf{Assistant}: 

\end{mdframed}
\newpage
\keypoint{Prompt for multi-granular fine-grained category generation}

\begin{mdframed}[backgroundcolor=gray!20]
\setlength{\parindent}{0pt}
\textbf{System: } You are a helpful assistant who can help with analyze image. 
    You will be provided with an image containing certain objects and a set of object classes.
    Your task is to generate corresponding specifc-level class names to the given classes.
    Important: the number of abstract classes you generated should be strictly same to the number of input classes.
    3-Layer Hierarchical Structure:\newline
        1. Fine-grained Level:\newline
            Combines the specific instance and subcategory levels.\newline
            Example: "Corgi," "Macbook," "SUV"\newline
        2. General Category Level:\newline
            Standard categories of objects.\newline
            Example: "Dog," "Computer," "Car"\newline
        3. Abstract Level:\newline
            Broad, overarching categories.\newline
            Example: "Animal," "Electronic Device," "Transportation"
    \newline\newline
    Few-shot Examples:
    \newline\newline
    **Image 1\newline
    - Given objects: fish, dog\newline
    - fine-grained labels: ["gold fish", "corgi"]
    \newline\newline
    **Image 2\newline
    - Present objects: laptop, container, box\newline
    - {{"fine-grained labels": ["Macbook", "mug", "mailer box"]}}
    \newline\newline
    **Image 3\newline
    - Present objects: Car, Road, Tree\newline
    - {{"fine-grained labels": ["Tesla", "Highway", "Oak"]}}
    \newline\newline
    \textbf{User:} Now for the following image, generate {granular}-level object class names for the input. 
    Generate a Json dictionary in the format of {{"fine-grained labels": ["the generated fine-grained labels class name", ...]}}. Only output the Json dictionary. Strictly no other output.
    Important: If there're multiple objects belong to the same abstract, be sure to give them the same abstract-level names. Note that the specific level object must present in the image. If you are not sure about the specific level or cannot distinguish, just output {{"fine-grained labels": "Unknown"}} only.

    **Query Image**  \newline
    - Present objects: [objects]
    
\textbf{Assistant}: 
\newpage
\end{mdframed}
\keypoint{Prompt for reasoning subset generation}
\begin{mdframed}[backgroundcolor=gray!20]
\setlength{\parindent}{0pt}
\textbf{System: } You are an intelligent chatbot designed to generate question-answer pairs according to the given image and a list of objects, each describing an object in the image you are observing. Your task is to return a question-answer pair where the question requires reasoning and the answer can correctly answer the question.

    The provided image has a height of [height] and a width of [width]. The image, image caption, objects in the image, and their respective bounding box coordinates are as follows:\par
    [**image**]
    \newline
    object at [bounding boxes]...
    \newline
    Coordinates represent (top-left x, top-left y, bottom-right x, bottom-right y). 
    \newline
    The question must be framed to require image reasoning for a response. Additional requirements for the generated question include:
    \newline
    1. The answer must reference the given object class or its equivalent and should not imply other potential objects.\newline
    2. The question should require some reasoning to answer and cannot be too broad.\newline
    3. The question should describe a activity or incorporate world knowledge.\newline
    4. You will output two forms of answers:\newline
        (1) objects: it contains the objects required to answer the question, each object should be in a form of object at [bounding box].\newline
        (2) text answer: it should be one or a few coherent sentences connecting the objects in (1) that answer the question. In the sentence, [seg-X] should be used to indicate the objects in (1), and X corresponds to the No.X objects in the list (0-indexed).
    \newline\newline
    Caption: a comfy room with table and sofa, and there is a laptop on the table.\newline
    Objects: \newline
        table at [120, 50, 240, 90];\newline
        pizza at [260, 80, 275, 100];\newline
        fork at [160, 30, 170, 35];\newline
        Plant at [315, 208, 320, 245]\newline
    Question: What steps do I need to take if I want to enjoy my meal?\newline
    Object answer: ["pizza at [160, 30, 180, 50]", "table at [120, 50, 240, 90]", "fork at [160, 30, 170, 35]"]\newline
    Text answer: "You can pick up the metal fork [seg-2] on the wooden table [seg-1] to slice the pizza [seg-0] and then enjoy."\newline
    \newline\newline
    **Image 2:**\newline
    Caption: a crowded road with bus and taxi.\newline
    Objects: \newline
        person at [335.52, 359.7, 347.85999999999996, 387.71];\newline
        boat at [285, 90, 337, 154];\newline
        person at [353.84, 390.06, 371.75, 412.82];\newline
        tree at [387.85, 3.42, 415.88, 40.67];\newline
        dog at [215, 362, 225.475, 381.84];\newline
    Question: What are people taking to reach the other side and how many people are there?\newline
    Object answer: ["boat at [285, 90, 337, 154]", "person at [335.52, 359.7, 347.85999999999996, 387.71]", "person at [353.84, 390.06, 371.75, 412.82]"]\newline
    Text answer: They are taking a boat [seg-0] to go to the other side of the river, and there are two persons [seg-1] [seg-2].\newline
    \newline
    **Query Image**\newline
    Question: {question}\newline
    Answer: {answer}\newline
    \newline\newline
    \textbf{User:} Return the generated question and answer in a Json dictionary only.

    **Query Image**  \newline
    - Present objects: [objects]...
    
\textbf{Assistant}: 

\end{mdframed}

\end{document}